\pdfoutput=1

\documentclass[11pt]{article}

\usepackage[final]{acl}

\usepackage{times}
\usepackage{latexsym}

\usepackage[T1]{fontenc}

\usepackage[utf8]{inputenc}

\usepackage{hyperref}       
\usepackage{url}            
\usepackage{booktabs}       
\usepackage{amsfonts}       
\usepackage{nicefrac}       
\usepackage{xcolor}         
\usepackage{caption}
\usepackage{subcaption}

\usepackage{amsmath}
\usepackage{amssymb}
\usepackage{mathtools}
\usepackage{amsthm}
\usepackage{mathrsfs}
\usepackage{graphicx}
\usepackage{bbm}
\usepackage{multirow}
\usepackage{makecell}
\usepackage{booktabs}
\usepackage{colortbl}
\usepackage{marvosym}
\usepackage{wasysym}
\usepackage{marvosym}
\usepackage[symbol]{footmisc}
\usepackage[skins,breakable]{tcolorbox}
\usepackage{nicematrix}
\usepackage{pifont}

\usepackage{microtype}

\usepackage{inconsolata}

\usepackage{graphicx}

\title{Rationales Are Not Silver Bullets: \\ Measuring the Impact of Rationales on Model Performance and Reliability}

\author{{\bf Chiwei Zhu\textsuperscript{\rm 1}, Benfeng Xu\textsuperscript{\rm 1,2\S}, An Yang\textsuperscript{\rm 2}, Junyang Lin\textsuperscript{\rm 2}} \\
{\bf Quan Wang\textsuperscript{\rm 3}, Chang Zhou\textsuperscript{\rm 2\textdagger}, Zhendong Mao\textsuperscript{\rm 1}} \\
\textsuperscript{1}University of Science and Technology of China \\ 
\textsuperscript{2}Qwen Team, Alibaba Inc \\ 
\textsuperscript{3}Beijing University of Posts and Telecommunications \\ 
\texttt{\{tanz, benfeng\}@mail.ustc.edu.cn}}

\begin{document}
\maketitle
\begin{abstract}
Training language models with rationales augmentation has been shown to be beneficial in many existing works.
In this paper, we identify that such a prevailing view does not hold consistently.
We conduct comprehensive investigations to thoroughly inspect the impact of rationales on model performance as well as a novel perspective of model reliability.
The results lead to several key findings that add new insights upon existing understandings: 1) Rationales can, at times, deteriorate model performance; 2) Rationales can, at times, improve model reliability, even outperforming their untrained counterparts; 3) A linear correspondence exists in between the performance and reliability improvements, while both are driven by the intrinsic difficulty of the task. These findings provide informative regulations on the broad utilization of rationales and raise critical implications on the procedure of explicitly aligning language models with implicit human thoughts. Codes can be found at \url{https://github.com/Ignoramus0817/rationales}.
\end{abstract}

\section{Introduction}
\label{sec:intro}
It is widely acknowledged that the capabilities of large language models can be significantly enhanced when they are given time to \textit{think}, a process where a \textbf{rationale} is generated first to mimic human inner thought before arriving at the final answer~\citep{wei2023chainofthought, kojima2023large}.
Although this concept was initially identified and established in the context of prompting large language models (LLMs), it has also been extensively explored in training language models (LMs) as well.
In general, rationales have profoundly influenced our understanding and utilization of LMs.
\footnotetext[4]{Corresponding author. Work done during the internship at Alibaba Group.}
\footnotetext[2]{Work done while working at Alibaba Group.}

\begin{figure}[t]
\centering
\includegraphics[width=\linewidth]{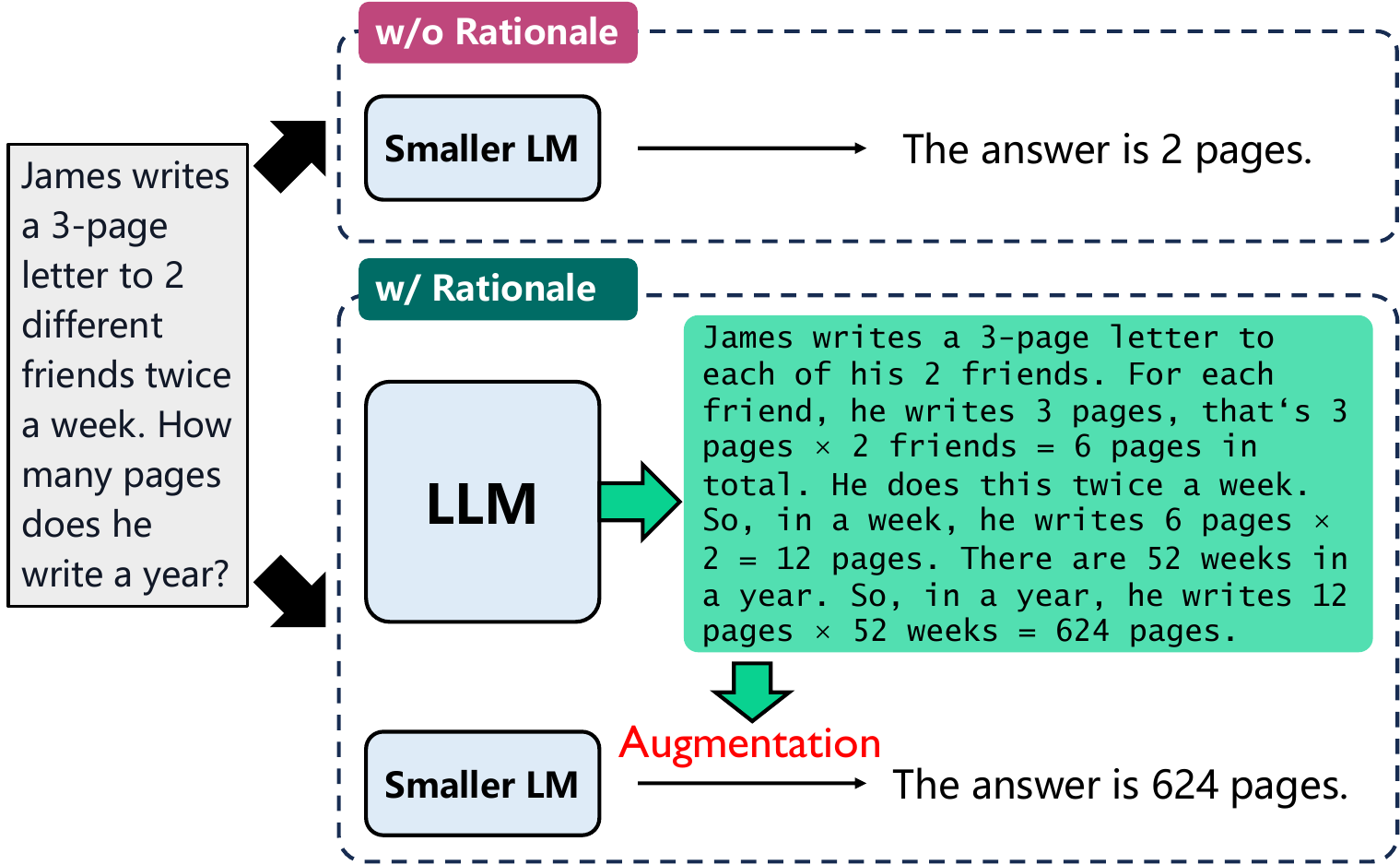}
\caption{Illustration of training LMs with rationale augmentation.}
\label{fig:rationales_demo}
\end{figure}

Benefiting from the powerful capabilities of LLMs~\citep{openai2023gpt4}, it is now much more approachable to obtain high-quality, large-scale synthetic reasoning traces as rationales.
As a result, \textbf{Rationale-Augmented finetuning (RAFT)} has been receiving increasing attention in many recent works, where a rationale is concatenated between the original question and answer to augment the learning process of a weaker or smaller language model (Figure~\ref{fig:rationales_demo}).
RAFT has brought consistent benefits for diversified tasks including mathematical reasoning~\citep{shridhar2023distilling, magister2023teaching}, question answering~\citep{wang2023pinto, li2022explanations}, symbolic reasoning~\citep{magister2023teaching} as well as general-purpose chatbots~\citep{li2023symbolic, mitra2023orca}.
As a result, adding rationales is becoming a default measure when finetuning smaller models recently.

In this paper, we identify some dissonance in the broad beneficial effect of RAFT, where introducing rationales can have a negative impact on model performance on certain tasks.
Through far more comprehensive and extensive investigations, we present results that diverge from previous expectations.
Aside from model performance, we include \textbf{Calibration} as an essential supplemental perspective of our analysis.
Calibration refers to whether a model can assign its predictions with proper probabilities that reflect the actual likelihood of its results being correct~\citep{guo2017calibration}.
It indicates the reliability of a deep model and can be affected when the prediction process is augmented with rationales.

\begin{figure}[t]
\centering
\includegraphics[width=\linewidth]{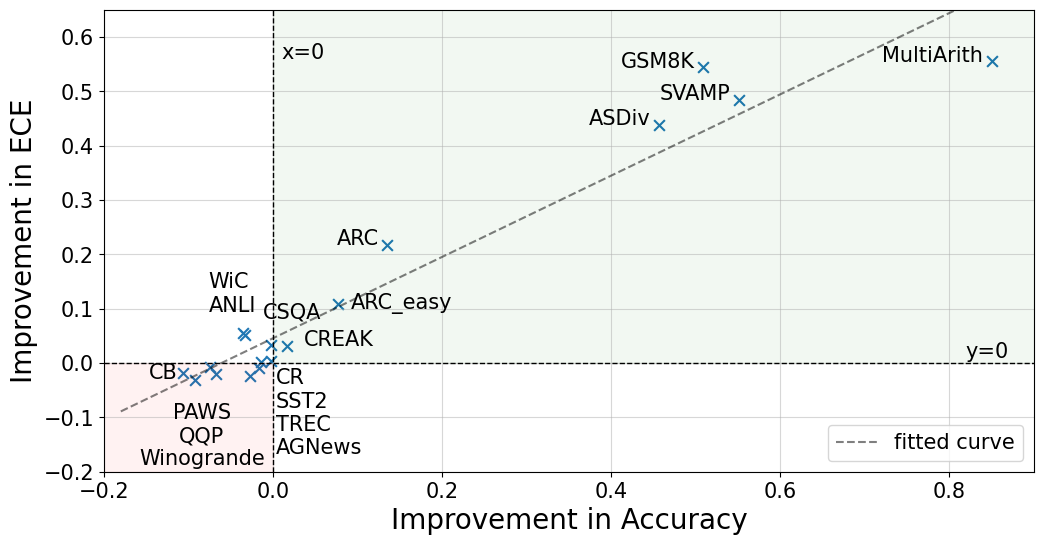}
\caption{Improvement in Accuracy ($x$-axis) and Expected Calibration Error (ECE) ($y$-axis) under RAFT for different datasets.}
\label{fig:acc_ece}
\end{figure}

Collectively, we measure the impact of RAFT on model performance and reliability on a total of 18 tasks. Figure~\ref{fig:acc_ece} provides an overview of all results, and key findings can be summarized as follows.

\definecolor{babypink}{rgb}{0.96, 0.76, 0.76}
\definecolor{brilliantlavender}{rgb}{0.96, 0.73, 1.0}
\definecolor{bubblegum}{rgb}{0.99, 0.76, 0.8}
\definecolor{brightube}{rgb}{0.82, 0.62, 0.91}
\definecolor{darkpastelpurple}{rgb}{0.59, 0.44, 0.84}
\definecolor{deeplilac}{rgb}{0.6, 0.33, 0.73}
\definecolor{grayblue}{rgb}{0.54, 0.60, 0.69}
\definecolor{riceyellow}{rgb}{0.75, 0.70, 0.62}
\begin{tcolorbox}[colback=black!5!white,colframe=grayblue!75!grayblue, boxsep=0mm, adjusted title=Key Findings I]
\textbf{On Performance:} Rationales indeed bring variable benefits for many difficult tasks, but this does not invariably hold across all tasks (corresponding to area \(x < 0\)).
\end{tcolorbox}
In 11 out of 18 tasks, performance has dropped when augmented with rationales, which is beyond expectation: one would expect RAFT to at least do no harm if it does not bring much improvements. It might be noticed that recent \texttt{gpt-4-o1} model employs rationales for better performance, seemingly challenging this work. Actually, it helps strengthen our conclusions. On the one hand, capabilities of \texttt{gpt-4-o1} mainly exhibits in math and code reasoning fields, which is consistent with our observations. On the other hand, it demonstrates that more sophisticated effort should be made to make better use of rationales.

\begin{tcolorbox}[colback=black!5!white,colframe=grayblue!75!grayblue, boxsep=0mm, adjusted title=Key Findings II]
\textbf{On Reliability:} Model calibration error can benefit from rationales, and even outperforms its untrained base model in certain tasks ($y>0$).
\end{tcolorbox}
Studies have pointed out that pretrained language models are well calibrated enough and finetuning process would degrade the calibration of language models~\citep{guo2017calibration, kadavath2022language, he2023preserving, zhu-etal-2023-calibration}.
However, when such a finetuning process is augmented with rationales, this degradation can be alleviated (12 out of 18 tasks). There are 3 tasks where calibration error under RAFT is even slightly better than its untrained base model.

\begin{tcolorbox}[colback=black!5!white, colframe=grayblue!75!grayblue, boxsep=0mm, adjusted title=Key Findings III]
\textbf{Performance-Reliability Correlation:} There exists a significant linear correspondence between the improvement in model performance and reliability under rationale-augmented training ($y=\alpha \cdot x + \beta$).
\end{tcolorbox}
We empirically find that rationale's impacts on performance and reliability are synchronized.
We attribute this linear correlation to the intrinsic difficulty of specific tasks. We further propose several difficulty metrics to validate this assumption and establish, for the first time, a quantitative relationship between correlation and task difficulty.

We also design extensive ablations to verify the robustness of our findings across varied conditions.
Finally, we locate the reason of the impacts of rationales and provide explanations with qualitative study. In appendices we also present exploratory analysis and discussions, delving deeper into the underlying mechanism of rationales and their impacts.
In general, this paper depicts a systematic view of the impacts of rationale-augmented finetuning, revealing a deeper understanding as well as new insights into its utilization and mechanisms. 

\section{Preliminaries}
\subsection{Rationale-Augmented finetuning}
\paragraph{Rationales.} Rationales are free-text reasoning steps produced by either human beings or language models when solving problems. Specifically, for language models, rationales can be defined as reasoning steps produced ahead of answers in Chain-of-Thought inference (see Fig.~\ref{fig:rationales_demo}).

\paragraph{Rationale-Augmented Finetuning.} RAFT is introducing rationales in the finetuning process. We conduct RAFT in the following steps: Given a supervised dataset $D=\{x_i, y_i\}_{i=1}^n$ where $x_i$ and $y_i$ are the $i_{th}$ input and label, rationales $r_i$ are first generated for each sample, resulting in the augmented dataset $D^\prime=\{x_i, (r_i, y_i)\}$. Then we finetune a language model on the augmented dataset maximizing the probability of generating $(r_i;y_i)$, where ``$;$" means concatenating. For comparison, we train the model using the same hyper-parameters but with solely answer labels, i.e. maximizing the probability of generating $y_i$. Then we compare the performance and calibration of the two models. To measure model performance and calibration quantitatively, we use \textbf{Accuracy (Acc)} and \textbf{Expected Calibration Error (ECE)} as metrics following previous works~\citep{guo2017calibration,li2023symbolic}.

\begin{figure*}[t]
\centering
\includegraphics[width=0.9\linewidth]{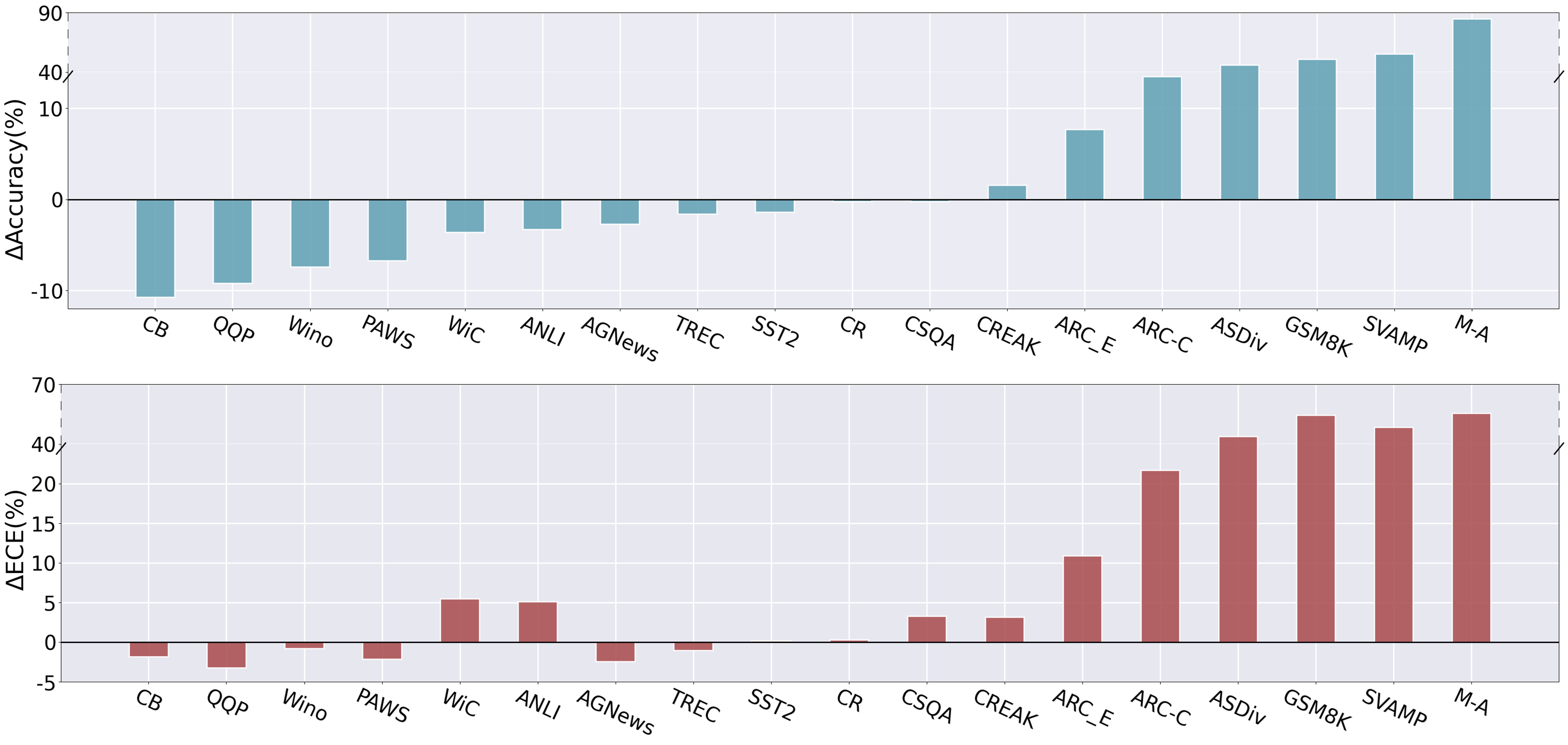}
\caption{Improvements in accuracy and ECE under RAFT. Datasets are re-ordered according to the improvements in accuracy. Wino refers to Winogrande and M-A refers to MultiArith. Note that y-axes are folded for better display. }
\label{fig:bar_acc_ece}
\end{figure*}

\subsection{Calibration}
\paragraph{Confidence Calibration.}
Given a classification problem where we have the input $X\in\mathcal{X}$, label $Y\in\mathcal{Y}=\{1,2...K\}$, which are random variables following ground truth joint distribution. Also we have model prediction $Y^{\prime}\in\mathcal{Y}$ and assigned confidence $P^{\prime}\in[0,1]$. A model is perfectly calibrated if it satisfies the following equation:
\begin{equation}
\label{eq_1}
P(Y^{\prime}=Y|P^{\prime}=p)=p,\quad\forall{p} \in{[0,1]}
\end{equation}
However, the probability in Equation~\ref{eq_1} is not calculable. To measure model calibration statistical approximations are often used, including Expected Calibration Error~\citep{naeini2015obtaining}.

\paragraph{Expected Calibration Error.}
Expected Calibration Error (ECE) is a quantitative measurement of calibration using finite data samples. Given a set of $N$ predictions and their confidence, we divide the confidence interval $[0,1]$ into $M$ bins with equal length $1/M$ and group these predictions according to their confidence. Then we can calculate the average accuracy and confidence of each bin:
\begin{equation}
\label{eq_2}
Acc(B_m)=\frac{1}{|B_m|}\sum_{i\in{B_m}}{\mathbbm{1}(\hat{y_i}=y_i)},
\end{equation}
\begin{equation}
\label{eq_3}
Conf(B_m)=\frac{1}{|B_m|}\sum_{i\in{B_m}}{\hat{p_i}},
\end{equation}
where $B_m$ is the set of indices of samples in the $m_{th}$ bin. $\hat{y_i}$ and $y_i$ are prediction and ground truth for the $i_{th}$ sample. $\mathbbm{1}$ is an indicator function generating 1 if the prediction is correct and 0 otherwise. $\hat{p_i}$ is the confidence that the model assigns to the $i_{th}$ prediction. Then ECE is calculated as:
\begin{equation}
\label{eq_4}
ECE=\sum_{m=1}^{M}{\frac{|B_m|}{N}|Acc(B_m)-Conf(B_m)|}
\end{equation}
ECE measures the gap between average confidence and accuracy among all bins. Lower ECE means better calibration. We use $M=10$ in all experiments, following previous works~\citep{guo2017calibration, desai2020calibration, he2023preserving}.

\subsection{Datasets and Implement Details}
\label{sec:implement}
\paragraph{Datasets} We conduct our research over 18 datasets in 7 categories, including Math Reasoning, Common Sense Reasoning, Sentiment and Topic Analysis, Paraphrasing, Natural Language Inference, Word Sense Disambiguation and Co-reference Resolution. Dataset details for each category can be found in Appendix \ref{app:datasets}.
\paragraph{Rationale Generation}
For each dataset, we construct a training set of 20k samples augmented with rationales. As obtaining large-scale human annotations is costly, we utilize \texttt{gpt-3.5-turbo-0613} to generate rationales following~\citep{fu2023specializing}. For each data point, we formulate the input with manually written prompts and query \texttt{gpt-3.5-turbo-0613} to generate rationales, from which we only keep rationales that lead to correct answers\footnote{For each sample we repeatedly generate until getting the correct answer or reaching the retry limit 10.}. We recursively traverse the dataset until we get enough data. Prompts for rationale generation are in Appendix \ref{app:prompts}. Though keeping only rationales leading to correct answers has already guaranteed the data quality to some extent, we further conduct a brief quality examination to make sure the generated rationales, which confirm the quality of the data(details in Appendix~\ref{app:quality_test}).

\begin{figure*}[t]
\centering
\includegraphics[width=0.8\textwidth]{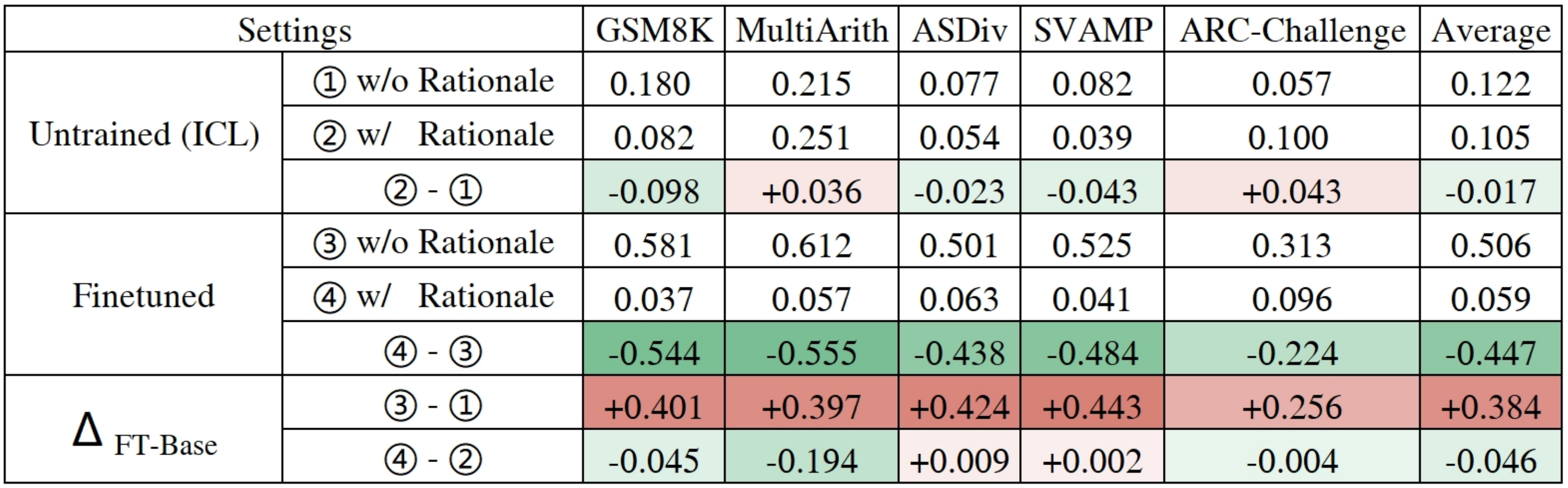}
\caption{ECE of finetuned and base models. $\Delta_{FT-Base}$ means difference of finetuned and pretrained models.}
\label{fig:ece_comparison}
\end{figure*}

\paragraph{Training and Inference.} We selected LLaMA-2-7B~\citep{touvron2023llama} as our base model. Full hyper-parameters for training are in Appendix \ref{app:hyper}. During inference, we apply a self-consistency voting~\citep{wang2023selfconsistency}. For every input question, we sample 10 reasoning paths with a temperature of 0.8, each generating an answer, from which the one that appears most frequently is kept as the final answer. If the answer appears $n$ times in the sampled results, we consider the confidence as $n/10$.

\section{Impacts of Rationales on Performance and Reliability}
We record the accuracy and ECE difference of models finetuned with and without rationales as follows:
\begin{equation}
    \Delta{Acc} = Acc_{RAFT} - Acc_{FT},
\end{equation}
\begin{equation}
    \Delta{ECE} = - (ECE_{RAFT} - ECE_{FT}),
\end{equation}
Note that we use a negative increase of ECE as lower ECE means better calibration.

\subsection{Impacts on Model Performance}
As is seen in Fig.~\ref{fig:bar_acc_ece}, the most significant improvement in accuracy happens in the math reasoning task, where accuracies of all four datasets improve by more than 40\%. Rationales also raise model performance on the two ARC datasets. Surprisingly, rationales do not always bring performance gain. Instead, they distract models from getting the correct answers for most of the tasks.

\subsection{Impacts on Model Reliability}
As can be seen in Fig.~\ref{fig:bar_acc_ece}, in most tasks models under RAFT are better calibrated than their counterparts finetuned with answer labels. Fig.~\ref{fig:ece_comparison} shows the results of tasks where model calibration improves the most. Firstly, from the table we can see that finetuning with labels does much harm to model calibration (\ding{174} - \ding{172}). Then it is noticeable that for both untrained and finetuned models, incorporating rationales brings benefit to model calibration (\ding{173} - \ding{172}, \ding{175} - \ding{174}). Lastly, in 3 tasks, models with RAFT can achieve even better ECE than the untrained models (\ding{175} - \ding{173}). Our results verify the established conclusion that finetuning will damage model calibration, while also showing that introducing rationales can alleviate such harmful effects.

\subsection{Linear Correlation Between Impacts on Performance and Reliability}
As can be seen in Fig.~\ref{fig:acc_ece}, the improvement in accuracy and ECE across different datasets show a significant linear correlation, depicted as follows:
\begin{equation}
    \Delta{ECE} = \alpha\Delta{Acc} + \beta,
\end{equation}
where $\alpha = 0.7479, \beta = 0.0456$. Line fitted with Ordinary Least Squares algorithm~\citep{galton1886regression} can be seen in Fig.~\ref{fig:acc_ece}. The Pearson Coefficient and p-value from the significance test are 0.9681 and $2.462e^{-10}$ respectively, portraying a very distinct linear correlation. Such correspondence may also explain why models become worse-calibrated for tasks where accuracy drops the most.

\begin{figure*}[ht]
\centering
\includegraphics[width=0.95\linewidth]{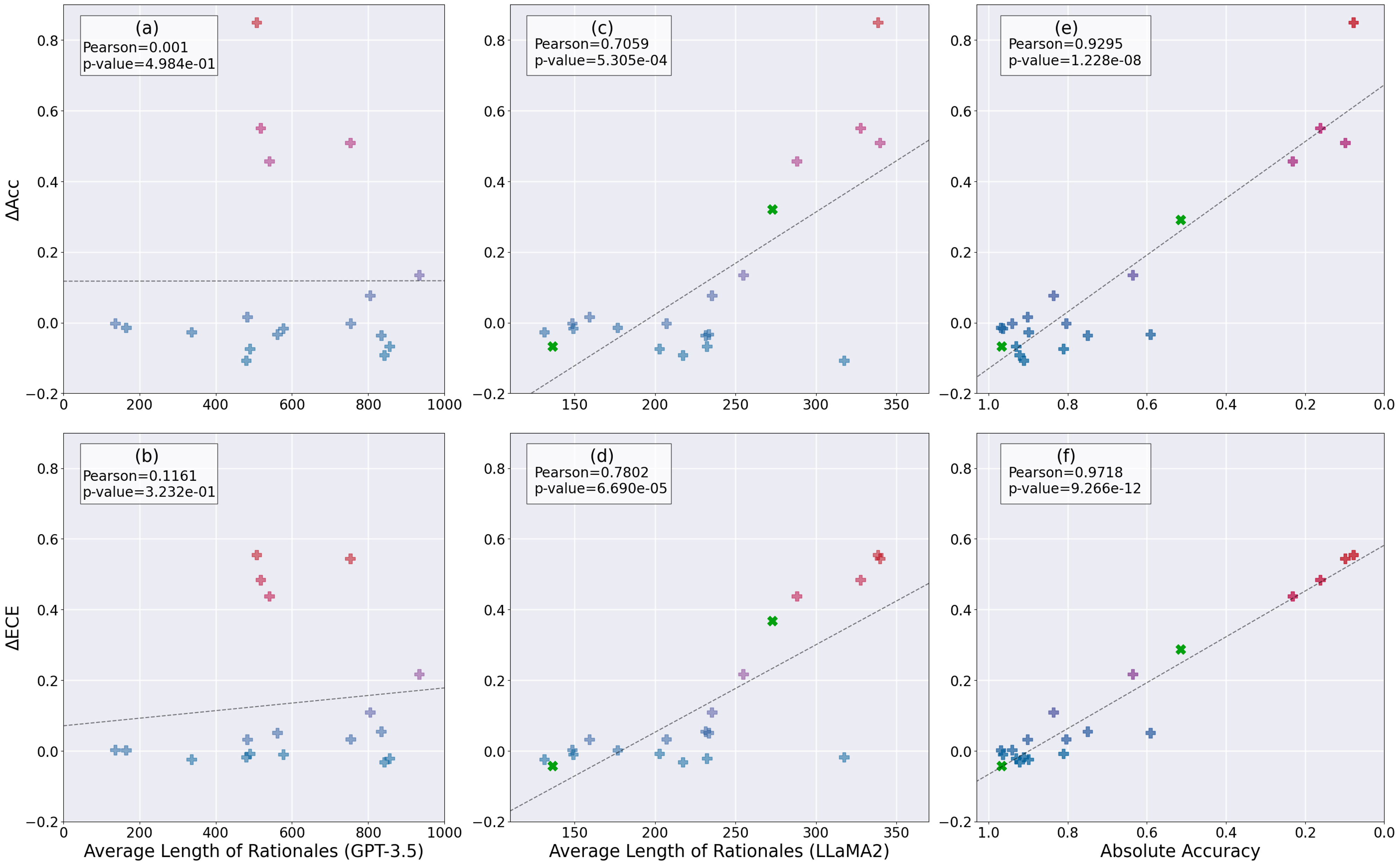}
\caption{Correspondence between model improvement and task difficulty across different metrics: (a), (b): Average length of rationales from \texttt{gpt-3.5-turbo-0613}. (c), (d): Average length of rationales from LLaMA-2-7B-base. (e), (f): Absolute Accuracy of LLaMA-2 model finetuned with answer labels (x-ticks is reversed). The color of the points shifts from blue to red as Y-values increase.} 
\label{fig:difficulty}
\end{figure*}

\section{Analyze Impacts of Rationales with Difficulty}
\label{sec:difficulty}
It is notable that improvements are more significant for intuitively harder tasks, e.g. ARC-Challenge compared to ARC-Easy. Thus we assume the impacts of rationales on performance and calibration are both related to the inherent difficulty of tasks, which can explain the linear correspondence. However, correctly defining the intrinsic difficulty of a task is non-trivial. In this section, we explore three alternative metrics to approach a reasonable characterization of task difficulty as follows.

\subsection{Definition of Task Difficulty}
\paragraph{Metric 1: Lengths of Rationales Generated by GPT-3.5.} 
It is well-acknowledged that the duration of human thought is positively correlated with problem difficulty~\citep{KOTOVSKY1985248, kahneman2011thinking}. We propose that a similar relationship exists for pretrained language models, as they possess extensive general world knowledge. This parallel between rationale length and computational effort is frequently drawn in contemporary research on inference-time scaling~\citep{snell2024scalingllmtesttimecompute}. As a further justification, we demonstrate on a subset that it shows distinct linear correlation with an established dataset difficulty metric, V-usable Information~\citep{ethayarajh2025understandingdatasetdifficultymathcalvusable}, with a Pearson coefficient of 0.895 (Details in Appendix~\ref{app:v_info}. Note that this method requires training a separate model for each dataset to measure, so we only use it for a preliminary study to demonstrate the effectiveness of our metrics).

So we use the average length of rationales generated by \texttt{gpt-3.5-turbo-0613} as a measurement for task difficulty. Specifically, for each test set, we sample 100 data points, apply a zero-shot Chain-of-Thought inference to each of them using \texttt{gpt-3.5-turbo-0613}, and collect the rationales. For each piece of data, we repeatedly generate 10 times, resulting in 1,000 rationales for the dataset, which we call \textbf{reference rationales} in the following parts. Then average length is calculated to measure the difficulty of this dataset.

\paragraph{Metric 2: Lengths of Rationales Generated by LLaMA2-base.}
As GPT-3.5 is trained with closed-source SFT data, the generated rationales might potentially suffer from certain biases. For instance, its SFT data may contain math questions similar to those in GSM8k, thus impacting the length of generated rationale as well as its neutrality as a difficulty metric.
In order to pursue unbiased rationales that naturally arises from pre-trained LLMs, we design a \textbf{Mix-of-Task In-Context Prompting} strategy.
We first employ LLaMA2 as the generator, since it is not explicitly aligned to human instructions, we prepend 3 demonstrations to regulate its behavior to generate rationales for each given instance.
The crucial aspect here is to make sure all three demonstration tasks are different from the target task so that the generated rationale is neither biased toward the alignment procedure nor biased towards any of the demonstrations.
The prompts are constructed as:
\begin{multline}
    P = \{(x_i,r_i,y_i)^{t_1};(x_j,r_j,y_j)^{t_2}; \\
    (x_k,r_k,y_k)^{t_3}; x^{t_n}\}
\end{multline}
where $(x_i,r_i,y_i)^{t_1}$ are input, rationale and answer sampled from task $t_1$. $x^{t_n}$ is the input from target task $t_n$. Demonstration tasks $t_1$, $t_2$, and $t_3$ are randomly sampled from the task pool excluding $t_n$.

\begin{figure}
    \centering
    \includegraphics[width=0.9\columnwidth]{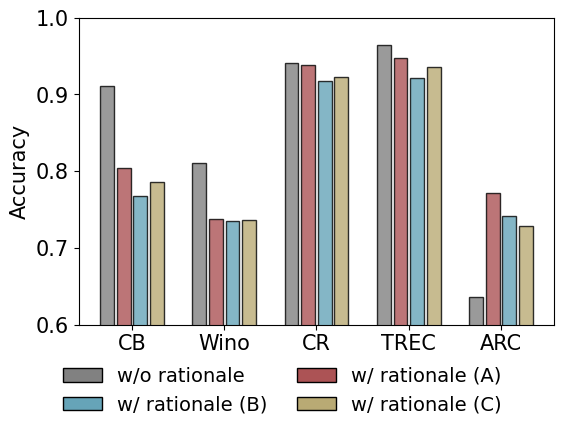}
    \caption{Impacts of varied rationales.}
    \label{fig:annotator}
\end{figure}

\paragraph{Metric 3: Absolute Accuracy.} Another alternative metric is the absolute accuracy of models. We assume that lower accuracy corresponds to higher task difficulty as models are less capable of these tasks. We consider the accuracy under vanilla label-only finetuning for consistency.

\subsection{Attributing Impacts of Rationales with Difficulty}
Fig.~\ref{fig:difficulty} shows how the improvement in accuracy and ECE evolves with the increase in task difficulty. Clear rising trends are evident in Fig.~\ref{fig:difficulty} (c), (d), (e), and (f), in which reference rationales generated by LLaMA-2 and absolute accuracy are used to measure difficulty. Meanwhile, we find the correlation between accuracy improvement and task difficulty bears a resemblance to that between ECE and task difficulty.
Such a trend may indicate that the improvements in accuracy and ECE are similarly driven by task difficulty, thus resulting in a linear correlation with each other. Note that when using GPT-3.5 generated rationales as references, there is no significant positive relation, which might be caused by the bias we mentioned before.

\subsection{Actionable Insight: Estimating Gain of RAFT with Task Difficulty}
The above findings in this paper directly points out an actionable insight: we can estimate the effects of RAFT based on various difficulty-related factors. Although strictly calculate the specific impacts is challenging, we manage to measure the difficulty of tasks with lengths of LLaMA2 generated rationales and absolute accuracy of models, which provides a preliminary solution and may serve as actionable guidance.
To be specific, the linear correspondence in Fig.~\ref{fig:difficulty} can be formulated as follows:
\begin{equation}
\Delta{Acc}=0.0029 \times Len(R_{LLaMA2}) -0.5567
\end{equation}
\begin{equation}
\Delta{Acc}=0.8031 \times Acc_{FT}+0.6730
\end{equation}
Based on the equations above, it is predictable with meaningful confidence whether and to what extent the performance and reliability improves when RAFT is applied for a given task. We verify the above relationship with two other datasets, SUBJ and CoinFlip (details in Appendix ~\ref{app:datasets}), which are green crosses in Figure~\ref{fig:difficulty}(c-f).
We believe our findings serve as a good initiation in the field of generating and using rationales more predictably.
Additionally, though this work mainly focuses on task-specific training, our conclusions may generalize to a broader topic, the alignment of large language models, which we discuss in Appendix~\ref{app:sft}.

\section{Robustness of the Impacts Brought by Rationales}
\label{sec:ablation}
As model performance and calibration can be affected by several factors, we conduct ablations to verify the robustness of our conclusions.
\paragraph{Model selection.} 
Extra experiments are conducted on Qwen-7B, LLaMA2-7B-Chat, and LLaMA2-13B. Additionally, we adopt another two stronger models as rationale generators, LLaMA-3.1-70B-Instruct and GPT-4o-mini (results in Appendix~\ref{app:models_hyper_para}). We perform RAFT on 6 datasets where performance varies in our previous experiments. As is shown in Fig.~\ref{fig:models} and Fig.~\ref{fig:teacher_models}, rationales still harm certain tasks across all models. And again, significant linear correspondence can be identified between the improvement in ECE and accuracy, despite that the slopes and intercepts are different. 
\paragraph{Hyper-parameters.} 
We conduct a search on learning rate (from \(5e^{-7}\) to \(5e^{-4}\)) and training epochs (from 2 to 6 epochs). Accuracy and ECE from the best models are reported (results in Appendix~\ref{app:models_hyper_para}). We select 3 datasets which are representative of the typical effect of RAFT: harm, improvement, or unchanging. In Table~\ref{tab:hyper_para} we can see that it remains unchanged whether RAFT brings improvement or harm to model performance and calibration on all datasets. Consequently, our conclusions about the impact of RAFT are robust against different models and hyper-parameters.
\paragraph{Different Prompts.} 
Rationales are generated with prompts from another two annotators (see Appendix \ref{app:other_annotators}) and are used for RAFT.
Results are displayed in Fig.~\ref{fig:annotator}. For all tasks we investigate, newly trained models behave the same way as the original ones, where RAFT uniformly improves or damages models' performance, which demonstrates the robustness of our key finding I. Key finding II is similarly verified consistent across different annotators in Appendix~\ref{app:ece_annotator}.
\paragraph{Multi-task Training.} 
We conduct multi-task training to verify our findings in a out-of-distribution scene. We utilize two methods of constructing training data for multi-task training:
\begin{itemize}
    \item Polarity Mixture: Separately mix data from datasets where RAFT brings gain (or harm).
    \item Full Mixture: Mix data from all datasets.
\end{itemize} 
We train models with mixed data and report test performance on QQP, CR, and GSM8K. Details and results can be found in Appendix~\ref{app:multi_task}. As shown in Fig.~\ref{fig:multi_task}, none of the multi-task models behaves differently from singlet-task models. As a result, multi-task instruction tuning does not change whether RAFT would cause improvement or harm.
\paragraph{Rationale-Augmented Prompting.}
We investigate whether these conclusions still hold under the frequently used inference paradigm, Rationale-Augmented Prompting.
We query \texttt{gpt-3.5-turbo-0613} to generate answers for the test sets with and without rationales. Still not all tasks benefit from rationales. However, the previous findings remain valid: incorporating rationales continues to improve ECE, with accuracy and ECE improvements showing a linear relationship. Finally, although there is a positive correlation between improvements and task difficulty, it is less pronounced than that observed in the RAFT settings (see Table \ref{table:linear_fit_comparison}), which may indicate that the effect of rationales is less influenced by task difficulty in the RAP setting. Full results are in Appendix \ref{app:rap}.

\definecolor{red1}{RGB}{190, 158, 150}
\definecolor{red2}{RGB}{197, 58, 50}
\definecolor{red3}{RGB}{237, 2, 126}
\begin{table*}[t]
\caption{Typical cases of different errors. Blue marks correct reasoning traces or golden answers and red marks oppositely.}
\label{table:case}
\begin{center}
\resizebox{0.9\linewidth}{!}{
\begin{tabular}{|c|c|c|m{10cm}|}
\hline
\multicolumn{2}{|c|}{Error Attribution} & Datasets & \multicolumn{1}{|c|}{Samples} \\
\hline
\multirow{8}{*}{\makecell{\textbf{\textcolor{teal}{Incorrect}} \\ \textbf{\textcolor{teal}{Reasoning}} \\(75\%)}} & \makecell{\textbf{\textcolor{teal}{Unreasonable}} \\ (57\%)} & Winogrande & \textbf{Question:} At night, Jeffrey always stays up later than Hunter to watch TV because \_ wakes up late. \newline \textbf{Answer:} ...Jeffrey stays up later to watch TV \textcolor{red}{because the other person wakes up late}. Therefore, the most likely option is \textcolor{red}{Hunter}. \textcolor{cyan}{ (Correct: Jeffrey)}\\
\cline{2-4}
& \makecell{\textbf{\textcolor{teal}{Off-Topic}} \\ (18\%)} & AGNews & \textbf{News:} Sun Microsystems Inc. has decided to postpone its September SunNetwork 2004 San Francisco user conference, and ... \newline \textbf{Answer:} ...The overarching topic of the news is \textcolor{red}{related to business, specifically the decisions and contemplations made by Sun Microsystems Inc.} regarding their conferences. The news is about \textcolor{red}{Business}. \textcolor{cyan}{ (Correct: Science/Technology)} \\
\hline
\multicolumn{2}{|c|}{\textbf{\textcolor{red2}{Trivial Reasoning}} (23\%)} & TREC & \textbf{Text:} What position did Willie Davis play in baseball ? \newline \textbf{Answer:} \textcolor{red}{This question is about the position played by Willie Davis in baseball}. So it is about an \textcolor{red}{Entity}. \textcolor{cyan}{ (Correct: Human Beings)}\\
\hline
\multicolumn{2}{|c|}{\textbf{\textcolor{red3}{Incorrect Conclusions}} (2\%)} & Winogrande & \textbf{Question:} Adam was law-abiding and not a crook like Joseph, so it was surprising that \_ was the one with a restraining order against him. \newline \textbf{Answer:} It implies that the person \textcolor{cyan}{who was law-abiding and not a crook like Joseph} is the one who had a restraining order against them. So the answer is \textcolor{red}{Joseph}. \textcolor{cyan}{ (Correct: Adam)} \\
\hline
\end{tabular}}
\end{center}
\end{table*}

\section{Explanatory Analysis}
In this section we try to explain the impacts observed above. We first exclude the formality of RAFT and the extra computation as potential causes through a blank rationale experiment and then conduct a qualitative analysis to identify situations where rationales cause performance drop. Furthermore, despite the difficulty of fully explaining our conclusions from the perspective of inner mechanisms of RAFT, we provide some hypotheses and a discussion in the Appendix~\ref{app:intrinsic_mechanism}.
\subsection{Locating the Cause of Impacts}
\label{sec:blank}
Given that rationales are additionally inserted between the question and answer, an intuitive argument would be that these reasoning traces should at least not deteriorate model performance if they do not provide obvious benefits.
We thus design an ablation experiment where rationales are replaced with equal-length blank sequence $\{\texttt{<Think\_i>}\}^L$ (See Appendix \ref{app:blank_rationales} for sequence example). 
This ablation can effectively disentangle the meaning and extra computation brought by rationales.

Results are shown in Appendix~\ref{app:blank_rationales}, where Fig.~\ref{fig:pads} shows that blank rationales indeed neutralize the negative impacts of rationales, bringing performance in line with label-only finetuning as expected.
We can thus exclude either the formality of RAFT or the extra computation as potential causes, and the real cause for performance deterioration should be the inherent meaning of the rationales. 

\subsection{Qualitative Analysis}
\label{sec:case_study}
To further investigate the performance drop, we collect 100 samples where models under RAFT generate wrong answers while models trained with labels do well. We conclude 3 types of errors as shown in Table \ref{table:case}.
\textbf{\texttt{\textcolor{teal}{Incorrect Rationales}}} refers to the rationales that directly lead to wrong answers. Specifically, \textbf{\texttt{\textcolor{teal}{Unreasonable}}} means the generated rationales are logically wrong. As shown by the example, the rationale is opposite to the meaning of the question. In such cases, models only mimic the form of reasoning steps but are not indeed reasoning. \textbf{\texttt{\textcolor{teal}{Off-topic}}} is a different situation, where the rationales make sense, but miss the key point to solve the problem. In the example, the rationale focuses on certain words, ignoring the topic itself. These cases are naturally difficult for models, where introducing rationales may not produce better results. \textbf{\texttt{\textcolor{red2}{Trivial Rationales}}} are those rationales that just repeat the problem. This usually happens for tasks with simple forms like sentiment/topic classification, where conclusions can be jumped to without much reasoning content. Finally, \textbf{\texttt{\textcolor{red3}{Incorrect Conclusions}}} are rare cases where generated answers are not consistent with the reasoning steps, which might be occasional and we do not further look into.

\section{Related Work}
\textbf{Rationales} are free-text reasoning steps produced by human or language models. Much attention has been paid to leveraging rationales to enhance model performance~\citep{zaidan-etal-2007-using, druck-etal-2009-active, zhang-etal-2016-rationale, camburu2018esnli}. Recently LLMs have been capable of generating reasoning steps through Chain-of-Thought~\citep{wei2023chainofthought, kojima2023large}, which inspires a bunch of works studying rationale-augmented finetuning (RAFT) using LLMs~\citep{nye2021work, chung2022scaling, fu2023specializing, shridhar2023distilling, mukherjee2023orca, mitra2023orca}. These works focus more on proposing new methods of extracting and utilizing rationales on certain target tasks. There are also works studying the effectiveness of rationales~\citep{carton2022learn, hase-bansal-2022-models, yao2023human, kabra2023programaided}.  Yet none of the above has studied the effect of rationales widely on different tasks nor do they study the effect of rationales on calibration (While \citep{sprague2024cot} introduces similar idea to ours, they mainly focuses on zero-shot setting, and do not study the behavior of calibration). Our study works in this untraveled direction.

\textbf{Confidence calibration} is first proposed for determining how a weather forecaster is reliable~\citep{miller1962statistical, murphy1973new}. Research on confidence calibration of statistical machine learning methods has a long history~\citep{degroot1983comparison, palmer2008toward}, and later calibration of neural networks is also researched on~\citep{nguyen2015posterior, hendrycks2016baseline, nixon2019measuring}.
~\citep{guo2017calibration} points out that finetuned neural networks are over-confident, which is potentially caused by over-minimizing the loss. Recently, calibration of language models also received attention.~\citep{desai2020calibration, kadavath2022language} study the calibration of pretrained language models and point out that they are well calibrated. Similarly, finetuning has been found to harm the calibration of LMs~\citep{he2023preserving, openai2023gpt4, zhu-etal-2023-calibration}. While previous works have studied the calibration of pretrained and finetuned language models, we expand the calibration measurement to the rationale-augmented finetuning setting.

\section{Conclusion}
In this work, we systematically examine the impact of rationale-augmented finetuning (RAFT) on model performance and reliability, and bring out several key findings that add new insights to current understandings. It sometimes deteriorates model performance, while alleviating calibration error caused by finetuning over-fitting. We also identify a significant linear correlation between the impacts on performance and reliability, both are driven by the intrinsic difficulty of the task.
With exploratory analysis and discussions, this paper implies future directions to continually delve into the underlying mechanism of rationales, pursuing better alignment between the explicit reasoning process of auto-regressive language models and implicit human thought structure.

\section*{Acknowledgment}
This research is supported by Artificial Intelligence-National Science and Technology Major Project 2023ZD0121200 and  National Natural Science Foundation of China under Grant 62222212.

\section*{Limitation}
\label{app:limitation_impact}
Our work reveals several new findings about rationale-augmented finetuning through extensive experiments. Although this work presents explanations and discussions about the intrinsic mechanism of rationales, theoretical proof is still pending. We hope this work would inspire new attempts on more rigorous formulation for rationale and its inherent mechanism on language models.

\section*{Potential Risks}
Our work focuses on the mechanism and effects of rationales, which provide guidance on the prevailing solution of rationale-augmented finetuning, and might also inspire improved methods to produce better synthetic rationales when training language models. Negative impact may include the abuse of LLMs to generate rationales for malicious tasks or using improved language models for harmful content generation.

\bibliography{custom}

\clearpage
\onecolumn
\appendix
\section{Further Discussion}
\subsection{Intrinsic Mechanism of Impacts of Rationales}
\label{app:intrinsic_mechanism}
In this section, we provide analysis and discussions that take a step further to explain our findings as well as unveil the intrinsic mechanism of rationales.
\paragraph{Why RAFT Causes Performance Drop?} 
We measure the information gain of a rationale as follows. 
Given a question $x$, rationale $r$ and answer $y$, in time step $t$ we construct an input $(x;r_{i<t})$, where $r_{i<t}$ are first $i$ tokens from the rationale. 
Then we query the model to generate 100 answers using this input. If $y$ appears $n$ times, we record $n/100$ as the probability of the model generating answer $y$. 
We repeat such step until the whole rationale is included.

\begin{figure}[h]
    \centering
    \includegraphics[width=0.6\columnwidth]{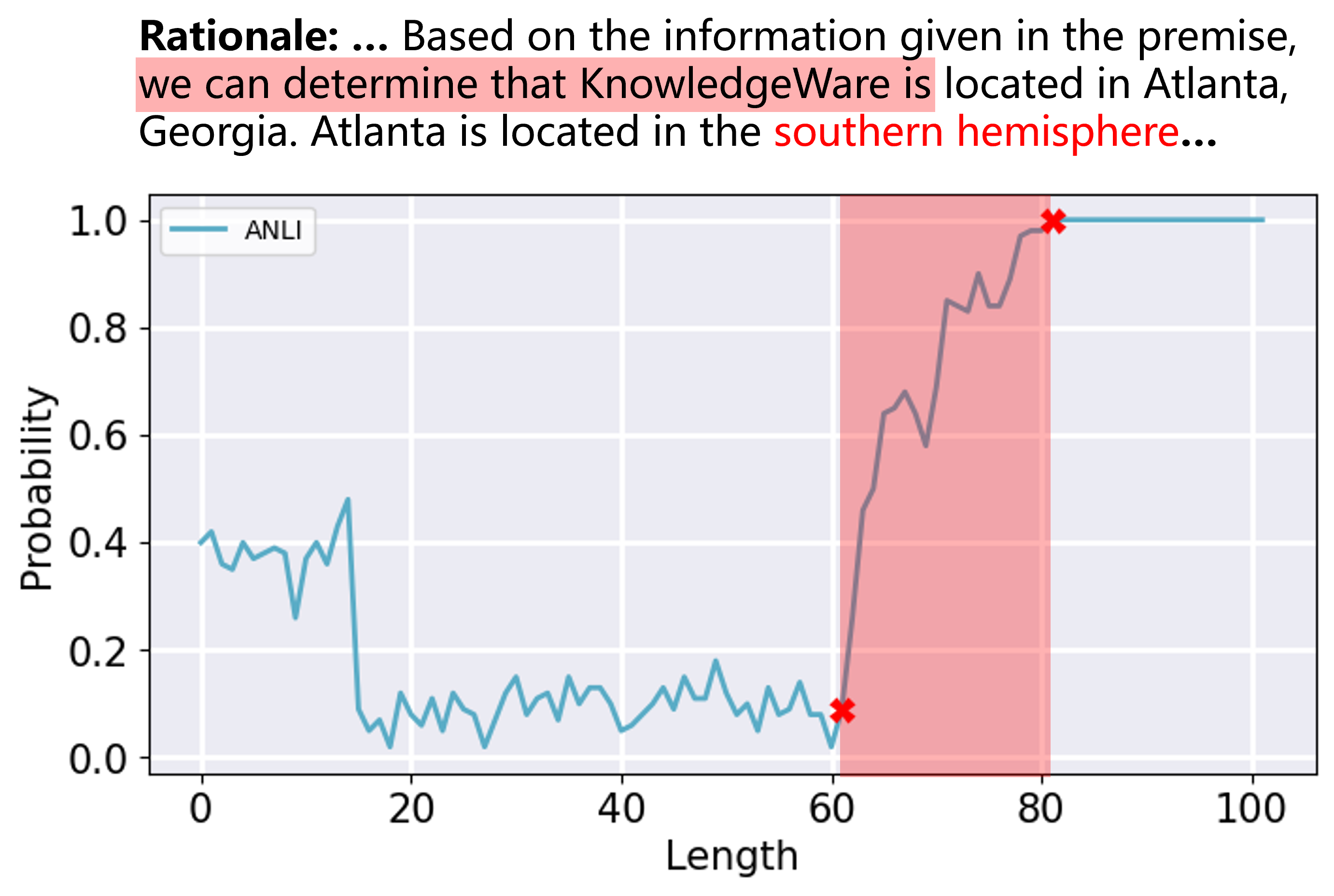}
    \caption{Probability of generating the final answer given different lengths of rationale tokens. Tokens corresponds to the sharp increase are highlighted in red.}
    \label{fig:information}
\end{figure}

Fig.~\ref{fig:information} shows how the probability of the final answer changes with the reasoning process. It fluctuates most time with only one sharp turn that leads to the final answer. This contradicts our expectation that rationales should decompose the task step-by-step, gradually bringing information gain and reducing uncertainty. This observation might be attributed to the inherent differences between human thoughts and LM architecture, it's difficult for LMs to fully mimic human thoughts when their capability is limited in an auto-regressive structure. As a result, it remains to be explored what is the golden rationale for an LM and how to construct it.

\paragraph{Why RAFT Benefit Model Calibration?} Previous study attributes the harmful effect of finetuning on calibration to the optimization process. In the optimization process, even when model predictions have already been correct, the loss can be further minimized by increasing the confidence of predictions~\citep{guo2017calibration}, which causes models to be over-confident and less calibrated. We suppose introducing rationales extends the labels from single numbers or letters to a longer text sequence space, which hinders the training process from minimizing loss by raising the confidence in final answers improperly. Besides, a widely used approach in RAFT is to include multiple rationales for one sample, which further strengthens such constraints.

\subsection{Possible Impacts on General LM Alignment}
\label{app:sft}
Though this work mainly studies task-specific finetuning, such conclusion might be generalized to a broader scenario, the alignment of large language models. The most applied technique for alignment for large language models is supervised finetuning (SFT), where human or model responses for a large amount of prompts are collected and used for finetuning models. In typical practice of SFT, each piece of data contains rationales, regardless of the task to which it pertains. At present it is rather unexplored whether we need rationale for all these prompts or how detailed rationales are proper for each prompt. According to our findings, rationales bring considerable income for some difficult tasks while helping less or distracting models for relatively simpler ones. Re-arranging the rationales in SFT data might enhance the instruction following ability of supervised finetuned models. An intuitive attempt is enhancing, impairing, or even removing the rationales for different prompts according to the task difficulty, which may serve as a refinement and denoising for SFT data. As it is not the main focus of this paper, we leave such exploration for future work.

\section{Details of Datasets and Licences.}
\subsection{Introduction of Datasets}
Here we introduce details about the datasets we used in the experiments. Note that for test sets whose gold answers are unavailable, we use the validation set instead. 
\label{app:datasets}
\paragraph{Math Reasoning.} MultiArith~\citep{roy-roth-2015-solving} is a collection of complicated arithmetic problems designed to test machine learning models. ASDiv~\citep{miao-etal-2020-diverse} is an elementary-school-level math word problem corpus that focuses on diversity when constructed. SVAMP~\citep{patel2021nlp} pays more attention to harder variations of basic problems as most benchmarks focus on difficult problems while models still lack the capabilities to deal with simpler ones. Lastly, GSM8K~\citep{cobbe2021training} is a collection of grade school math problems aimed at challenging the most advanced problem solvers.
\paragraph{Common Sense Reasoning.} ARC (AI2Reasoning Challenge)~\citep{Clark2018ThinkYH} and CommonSenseQA~\citep{talmor2019commonsenseqa}  are two common sense question answering datasets consisting of science exam questions and human written questions based on concepts in a common sense knowledge-base, CONCEPTNET~\citep{speer2018conceptnet}. CREAK~\citep{onoe2021creak} is a dataset combining common sense and entity knowledge, which includes inference and fact-checking questions for real-world or fictional entities (e.g. Harry Potter). 
\paragraph{Sentiment/Topic Analysis.} CR~\citep{ding2008holistic} are SST-2~\citep{socher-etal-2013-recursive} are the two datasets used for sentiment analysis, where each sample is labeled positive or negative. TREC~\citep{li2002learning} and AGNews~\citep{zhang2015character} are topic classification datasets consisting of questions and news snippets respectively. 
\paragraph{Paraphrase.} PAWS~\citep{zhang2019paws} and QQP\footnote{\url{https://quoradata.quora.com/First-Quora-Dataset-Release-Question-Pairs}} consists of sentence pairs from Wikipedia and Quora, which one should determine whether the sentences in the pair have the same semantic meaning.
\paragraph{Natural Language Inference (NLI).} CommitmentBank is a natural inference dataset from the SuperGLUE benchmark~\citep{wang2020superglue} and AdversarialNLI~\citep{nie2020adversarial} is an NLI dataset made up of adversarially constructed questions.
\paragraph{Word Sense Disambiguation.} WiC~\citep{pilehvar2019wic} is another sub-task of SuperGLUE, where one should determine whether a given word shows the same meaning in two sentences.
\paragraph{Coreference Resolution.}  We choose Winogrande~\citep{sakaguchi2019winogrande} in this task, where the task is to identify the specific entity a pronoun points to.
\paragraph{Additional Datasets for Validation} SUBJ~\citep{pang-lee-2004-sentimental} is a review analysis dataset, which mainly focuses on deciding the objectivity and subjectivity of reviews for films. CoinFlips~\cite{coin-flips} is a reasoning task where one should decide whether a coin is head or tail after a series of possible flips.

\subsection{Licenses of Data and Models}
\label{app:license}
We list all licenses of used data and models here except those whose licenses are not specified. We also cite all original papers for these assets.
\subsubsection{MIT}
SVAMP, GSM8K, CommonsenseQA, CREAK, CR, CoinFlips.
\subsubsection{CC}
ASDiv (CC BY-NC 4.0), ARC (CC BY-SA), ANLI (CC BY-NC 4.0), WiC (CC BY-NC 4.0).
\subsubsection{Apache-2.0}
Winogrande, Qwen models
\subsubsection{Custom Licenses}
AGNews (non-commercial use\footnote{\url{http://groups.di.unipi.it/~gulli/AG_corpus_of_news_articles.html}}) \\
PAWS (free use for any purpose\footnote{\url{https://github.com/google-research-datasets/paws/blob/master/LICENSE}}) \\
QQP (Quora ToS, non-commercial use\footnote{\url{https://www.quora.com/about/tos}}) \\
CommitmentBank (SuperGlue, research-only\footnote{\url{https://super.gluebenchmark.com/faq}}) \\
LLaMA2 (Llama 2 Community License Agreement\footnote{\url{https://huggingface.co/meta-llama/Llama-2-7b-chat-hf/blob/main/LICENSE.txt}})
\subsubsection{Unknown}
MultiArith, SST2, TREC, SUBJ

\clearpage
\section{Brief Quality Assessment for Generated Rationales}
\label{app:quality_test}
We sample 20 data points for each of the 18 datasets, resulting in a total 360 data points. We prompt GPT-4o with the following instruction to assess the helpfulness of each rationale.
\begin{tcolorbox}[colback=black!5!white,colframe=black!75!black,title=Prompt Templates for Rationales Quality Assessment, breakable]
Following you will see a question and the corresponding answers along with rationales generated by a language model. You should assess whether the reasoning steps in the rationales are helpful for getting the correct answer. Give your analysis first and conclude in the end with "Helpful" or "Unhelpful".
\end{tcolorbox}
In the evaluation result, only 1 of the rationales are labeled as "Unhelpful", which is then examined by human and turns out to be incorrectly classified by GPT-4o. Such results also demonstrate the quality of the rationales. Details of the "Unhelpful" sample are shown below:
\begin{tcolorbox}[colback=black!5!white,colframe=black!75!black,title=The False Negative Sample, breakable]
\textbf{Question:} \\
Decide whether the sentiment of the given statement is positive or negative: with mediasource being so easy , i found no use for the bundled nomad explorer software
\\\\
\textbf{Answer:}\\
Let's think step by step.\\
(1) Mediasource being easy to use suggests it is a positive aspect, making managing media files simple and efficient.\\
(2) The statement mentions finding no use for the bundled Nomad Explorer software. This implies that the software was not needed as the mediasource is already easy enough, indicating a negative experience.\\
Though there are positive parts in the sentence, The sentiment of the statement leans towards being negative."""
\\\\
\textbf{GPT-Evaluation:}\\
1. The rationale correctly identifies that the statement describes Mediasource as easy to use, which implies a positive experience with this tool.\\
2. The statement mentions that the user found \"no use for the bundled Nomad Explorer software.\" This does not necessarily indicate a negative experience with Nomad Explorer itself. It could simply mean that Mediasource was sufficient for the user's needs, rendering Nomad Explorer redundant. The rationale incorrectly interprets this as a negative sentiment.\\
...Unhelpful
\\\\
\textbf{Human-Evaluation:}\\
Helpful
\end{tcolorbox}

\clearpage
\section{Prompts Used in Rationales Generation.}
\label{app:prompts}
\subsection{Prompt Templates for Rationale Generation}
Here are the prompt templates for rationale generation, where red texts are fields from datasets and blue texts are content generated by LLM.
\begin{tcolorbox}[colback=black!5!white,colframe=black!75!black,title=Prompt Templates for Rationales Generation (Annotator A), breakable]
\lbrack \textbf{System Prompt}\rbrack\\
\texttt{You are a helpful assistant.} \\
\\
\lbrack \textbf{Math Reasoning}\rbrack \\
\texttt{Solve the following math problems. Add a line "The answer is n" at the end where n is the answer value.} \\ \\
\texttt{Question: {\color{red}\{\text{question}\}}} \\ \\
\texttt{Answer: Let's think step by step. {\color{blue}\{\text{LLM output}\}}} \\
\\ \\
\lbrack \textbf{Natural Language Inference}\rbrack \\
\texttt{Following is a premise and a hypothesis, determine whether the hypothesis is entailed by the premise, contradictory with the premise, or can not be determined. Add a line "The answer is x" in the end where x is your choice.} \\ \\
\texttt{Premise: {\color{red}\{\text{premise}\}}} \\
\texttt{Hypothesis: {\color{red}\{\text{hypothesis}\}}} \\
\texttt{Options: (A) Entailment (B) Contradiction (C) Neutral} \\ \\
\texttt{Answer: Let's think step by step. {\color{blue}\{\text{LLM output}\}}} \\
\\ \\
\lbrack \textbf{Sentiment Analysis}\rbrack \\
\texttt{Follwing is a statement, determine whether the sentiment is negative or positive. Add a line "The answer is x" in the end where x is your choice.} \\ \\
\texttt{Statement: {\color{red}\{\text{statement}\}}} \\
\texttt{Options: (A) positive (B) negative} \\ \\
\texttt{Answer: Let's think step by step. {\color{blue}\{\text{LLM output}\}}} \\
\\ \\
\lbrack \textbf{Topic Classification (TREC)}\rbrack \\
\texttt{Following is a question, determine which topic it is about. Add a line "The answer is x" in the end where x is your choice.} \\ \\
\texttt{Question: {\color{red}\{\text{question}\}}} \\
\texttt{Options: (A) Description and abstract concept (B) Entity (C) Abbreviation (D) Human being (E) Location (F) Numeric value} \\ \\
\texttt{Answer: Let's think step by step. {\color{blue}\{\text{LLM output}\}}} \\
\\ \\
\lbrack \textbf{Topic Classification (AGNews)}\rbrack \\
\texttt{Following is a piece of news and its brief description, determine which topic it is about. Add a line "The answer is x" in the end where x is your choice.} \\ \\
\texttt{News: {\color{red}\{\text{news}\}}} \\
\texttt{Options: (A) World (B) Sports (C) Business (D) Science and Technology} \\ \\
\texttt{Answer: Let's think step by step. {\color{blue}\{\text{LLM output}\}}} \\
\\ \\
\lbrack \textbf{Common Sense (ARC\&CSQA)}\rbrack \\
\texttt{Following is a selective question and its answer options. Select the most possible one. Add a line "The answer is x" in the end where x is your choice.} \\ \\
\texttt{Question: {\color{red}\{\text{question}\}}} \\
\texttt{Options: {\color{red}\{\text{options}\}}} \\ \\
\texttt{Answer: Let's think step by step. {\color{blue}\{\text{LLM output}\}}} \\
\\ \\
\lbrack \textbf{Common Sense (CREAK)}\rbrack \\
\texttt{Following is a statement, determine whether is is true or false based on common sense and fact. Add a line "The answer is x" in the end where x is your choice.} \\ \\
\texttt{Statement: {\color{red}\{\text{statement}\}}} \\
\texttt{Options: (A) False (B) True} \\ \\
\texttt{Answer: Let's think step by step. {\color{blue}\{\text{LLM output}\}}} \\
\\ \\
\lbrack \textbf{Paraphrase}\rbrack \\
\texttt{Following are two similar sentences. Determine whether they are asking the same question or describing the same situation. Add a line "The answer is x" in the end where x is your choice.} \\ \\
\texttt{Sentence1: {\color{red}\{\text{sentence1}\}}} \\
\texttt{Sentence2: {\color{red}\{\text{sentence2}\}}} \\
\texttt{Options: (A) Different (B) Same} \\ \\
\texttt{Answer: Let's think step by step. {\color{blue}\{\text{LLM output}\}}} \\
\\ \\
\lbrack \textbf{Coreference Resolution}\rbrack \\
\texttt{Following is a sentence where a word is replaced with a blank symbol "\_". You will be given two options and you should choose the most possible one to fill in the blank. Add a line "The answer is x" in the end where x is your choice.} \\ \\
\texttt{Sentence: {\color{red}\{\text{sentence}\}}} \\
\texttt{Options: {\color{red}\{\text{options}\}}} \\ \\
\texttt{Answer: Let's think step by step. {\color{blue}\{\text{LLM output}\}}} \\
\lbrack \textbf{Word Sense Disambiguation}\rbrack \\
\texttt{Following is a target word and two sentences. Determine whether the words in the two sentences have the same semantic meaning. Add a line "The answer is x" in the end where x is your choice.} \\ \\
\texttt{Target word: {\color{red}\{\text{target word}\}}} \\
\texttt{Sentence1: {\color{red}\{\text{sentence1}\}}} \\
\texttt{Sentence2: {\color{red}\{\text{sentence2}\}}} \\
\texttt{Options: (A) Different (B) Same} \\ \\
\texttt{Answer: Let's think step by step. {\color{blue}\{\text{LLM output}\}}} \\
\end{tcolorbox}

\subsection{Prompt Templates for Different Lengths}
\label{app:promtp_length}
\begin{tcolorbox}[colback=black!5!white,colframe=black!75!black,title=Prompt Templates for Different Lengths]
\lbrack \textbf{1}\rbrack 
\texttt{...Give a thorough analysis of the problem and explain your solution.} \\
\\
\lbrack \textbf{2}\rbrack
\texttt{...Analyze the problems and explain your solution as detailed as possible.} \\
\\
\lbrack \textbf{3}\rbrack
\texttt{...Let's think step by step.} \\
\\
\lbrack \textbf{4}\rbrack
\texttt{...Explain your solution with a few words.} \\
\\
\lbrack \textbf{5}\rbrack 
\texttt{...Explain the solution as short as you can.}
\end{tcolorbox}

\subsection{Prompt Templates from Annotator B}
\label{app:other_annotators}
\begin{tcolorbox}[colback=black!5!white,colframe=teal!75!teal,title=Prompt Templates for Rationales Generation (Annotator B), breakable]
\lbrack \textbf{Natural Language Inference}\rbrack \\
\texttt{You are now required to perform a natural language inference task. I will provide you with a premise and a hypothesis, and you need to determine whether the hypothesis can be inferred from the premise.} \\
\\
\lbrack \textbf{Sentiment Analysis}\rbrack \\
\texttt{You are now required to perform a sentiment analysis task. I will give you a text passage, and you need to determine whether it is positive or negative.} \\
\\
\lbrack \textbf{Topic Classification}\rbrack \\
\texttt{You are now required to perform a topic classification task. I will give you a sentence, and you need to determine which of the six possible topics it belongs to.} \\
\\
\lbrack \textbf{Common Sense (ARC)}\rbrack \\
\texttt{You are now required to perform a coreference resolution task. I will give you a text passage in which I intentionally omit a word, and then provide you with two options. You need to determine which option is more fitting in the context.} \\
\\
\lbrack \textbf{Coreference Resolution}\rbrack \\
\texttt{You are now required to perform a coreference resolution task. I will give you a text passage in which I intentionally omit a word, and then provide you with two options. You need to determine which option is more fitting in the context.}
\end{tcolorbox}

\subsection{Prompt Templates from Annotator C}
\begin{tcolorbox}[colback=black!5!white,colframe=olive!75!olive,title=Prompt Templates for Rationales Generation (Annotator C), breakable]
\lbrack \textbf{Natural Language Inference}\rbrack \\
\texttt{Review a given premise, determine whether the relevant hypothesis can be logically inferred from the premise.} \\
\\
\lbrack \textbf{Sentiment Analysis}\rbrack \\
\texttt{You will see a passage of text, please determine the sentiment of the text.} \\
\\
\lbrack \textbf{Topic Classification}\rbrack \\
\texttt{Given a sentence, please determine which of the six categories it belongs to.} \\
\\
\lbrack \textbf{Common Sense (ARC)}\rbrack \\
\texttt{You will be shown a question and four answers, from which you need to choose one based on your common sense.} \\
\\
\lbrack \textbf{Coreference Resolution}\rbrack \\
\texttt{Given a passage of text where a word is left blank, and you will see two options, each corresponding to a word. Please infer, based on the context, which word is more suitable to fill in the blank.}
\end{tcolorbox} 

\clearpage
\section{Training Settings.}
\label{app:hyper}
Table~\ref{table:hyperparam} is the hyper-parameters used in finetuning models. All models are trained for 3 epochs using Huggingface Transformers Library\footnote{https://huggingface.co/docs/transformers/index} on 4 NVIDIA A-100 GPU and Fully Sharded Data Parallel (FSDP).

\begin{table}[h]
\caption{Hyper-parameters of finetuning.}
\label{table:hyperparam}
\begin{center}
\begin{small}
\begin{sc}
\begin{tabular}{lr}
\toprule
Parameters & Values \\
\midrule
Epoch      & 3 \\
Learning Rate & $5e^{-6}$ \\
Batch size per device & 4 \\
Gradient Accumulation & 4 \\
Gradient Checkpointing & True \\
Precision  & BF16 \\    
Max Length & 2048 \\
Warmup Ratio & 0.03 \\
Weight Decay & 0 \\
Learning Rate Scheduler & Cosine \\
\bottomrule
\end{tabular}
\end{sc}
\end{small}
\end{center}
\end{table}

Below are an overview of training and inference time.

\begin{table}[h]
\caption{Training/Inference time overview.}
\label{table:time_overview}
\begin{center}
\begin{small}
\begin{sc}
\begin{tabular}{ccc}
\toprule
Model & \multicolumn{2}{c}{Training Time (Mins)} \\
Scale & w/o rationales & w/ rationales \\
\midrule
7B & \textasciitilde25 & \textasciitilde40 \\
13B & \textasciitilde60 & \textasciitilde90 \\
\midrule
\multicolumn{3}{c}{Inference time (seconds per sample)} \\
\midrule
7B & \textasciitilde0.12 & \textasciitilde40 \\
13B & \textasciitilde3.5 & \textasciitilde60 \\
\bottomrule
\end{tabular}
\end{sc}
\end{small}
\end{center}
\end{table}

\section{Correlation between V-usable Information and Lengths of Rationales}
\label{app:v_info}
Table~\ref{table:v_info} is the V-usable Information and average length of rationales for a dataset in each task category, where a distinct linear correlation comes up.
\begin{table}[h]
\caption{The V-usable Information and average length of rationales for a dataset in each task category. We denote V-usable Information as V-INFO and average length of rationales as LENGTHS. The Pearson Coefficient is 0.895.}
\label{table:v_info}
\begin{center}
\begin{small}
\begin{sc}
\begin{tabular}{cccccccc}
\toprule
Metrics & SST2 &	ANLI &	PAWS &	WiC & Winogrande &	ARC Challenge &	GSM8K \\
\midrule
V-Info &	-0.222 &	-0.0013 &	0.0011 &	0.0023 &	0.0044 &	0.0014 &	0.178 \\
Lengths &	176.1 &	202.7 &	232.2 &	231.3 &	233.3 & 	254.8 &	339.7 \\
\bottomrule
\end{tabular}
\end{sc}
\end{small}
\end{center}
\end{table}

\newpage
\section{Results of Different Models and Hyper-parameters}
\label{app:models_hyper_para}
\subsection{Different Models}
\begin{figure*}[h]
\begin{center}
\includegraphics[width=0.8\textwidth]{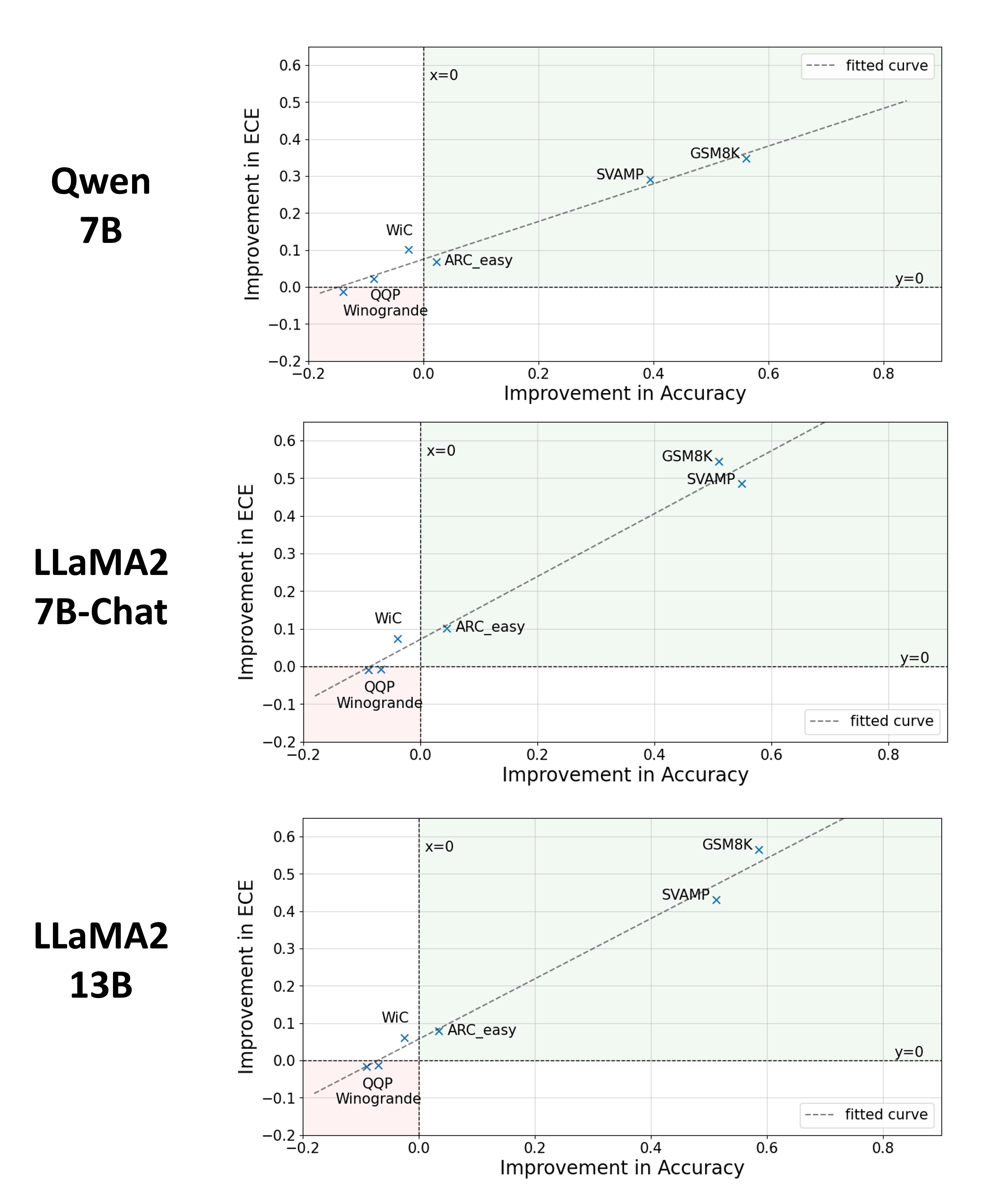}
\caption{Improvement in Accuracy and ECE of different models. Significant linear correspondence can still be observed.}
\label{fig:models}
\end{center}
\end{figure*}

\begin{figure*}[h]
\begin{center}
\includegraphics[width=0.8\textwidth]{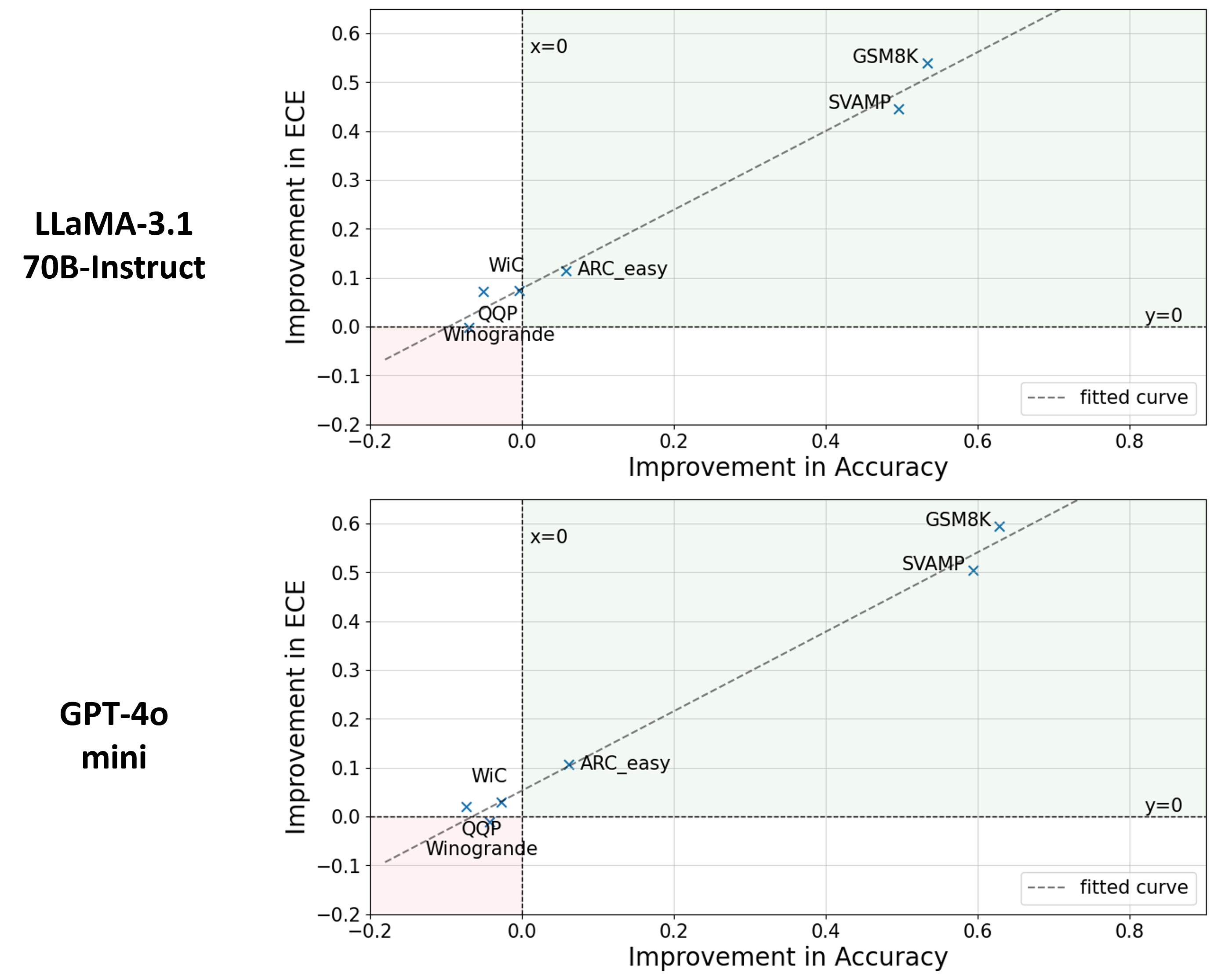}
\caption{Improvement in Accuracy and ECE of models trained on rationales generated with LLaMA-3.1-70B-Instruct and GPT-4o-mini. Significant linear correspondence can still be observed.}
\label{fig:teacher_models}
\end{center}
\end{figure*}

Fig.~\ref{fig:models} and Table~\ref{tab:models} are experimental results of different models and hyper-parameters. We can still observe significant linear correspondence between the improvement in accuracy and ECE in other models with high Pearson Coefficient and near-zero P-value. Besides, when models are trained with different hyper-parameters, our main conclusions still hold. Fig.~\ref{fig:teacher_models} shows results of LLaMA-2 fine-tuned on rationales generated with LLaMA-3.1-70B-Instruct and GPT-4o-mini. The conclusions are the same.

\begin{table}[h]
\caption{Significance test of linear correspondance in different models.}
\begin{center}
\begin{small}
\begin{tabular}{ccc}
\toprule
Models & Pearson & P-value \\
\midrule
Qwen-7B & 0.9883 & \(1.024\times10^{-4}\) \\
LLaMA2-7B-Chat & 0.9902 & \(7.248\times10^{-5}\) \\
LLaMA2-13B & 0.9943 & \(2.469\times10^{-5}\) \\
\bottomrule
\end{tabular}
\label{tab:models}
\end{small}
\end{center}
\end{table}

\clearpage
\subsection{Hyper-parameters}
Table.~\ref{tab:hyper_para} shows the results from the best models in the parameter search. For comparison, we also list results of our main experiments. As can be seen in the table, it remains unchanged whether RAFT brings improvement or harm to model performance and calibration.

\begin{table*}[h]
\caption{Results from best models in the parameter search.}
\begin{center}
\begin{small}
\begin{tabular}{ccccc}
\toprule
Metric &  \(\Delta{Acc}\) (paper) & \(\Delta{Acc}\) (w/ parameter search) &	\(\Delta{ECE}\) (paper) & \(\Delta{ECE}\) (w/ parameter search) \\
\midrule
QQP & -0.092 & -0.047 & -0.032 & -0.004 \\
CR &  -0.002 & -0.006 & 0.003 & 0.01\\
GSM8K & 0.509 & 0.509 & 0.544 & 0.394 \\
\bottomrule
\end{tabular}
\label{tab:hyper_para}
\end{small}
\end{center}
\end{table*}

\section{ECE Results Produced with Different Prompts}
\label{app:ece_annotator}
Following are ECE results of models trained with rationales produced with different prompts. We can see that difference of ECE is minor among models trained with different rationales and our conclusion consistently holds.

\begin{figure}[h]
\begin{center}
\includegraphics[width=0.5\columnwidth]{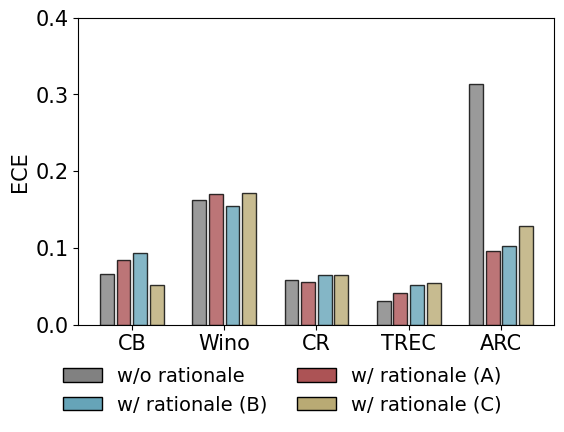}
\caption{ECE of models trained with rationales produced by annotator B and C.}
\end{center}
\end{figure}

\clearpage
\section{Details of Blank Rationales}
Here are experimental details of Section \ref{sec:blank}. We conduct experiments on three datasets, WiC, QQP and Winogrande. For each sample, we substitute the rationale with a blank sequence composed of \texttt{<Think\_n>} to ensure that it does not provide any additional information, as is shown in Table~\ref{XXX}. The number of blank \texttt{think} tokens is set as the average length of rationales in the dataset, specific numbers are listed in Table~\ref{tab:blank_detail}.

\begin{table}[h]
\caption{Length of blank rationales for each dataset.}
\begin{center}
\begin{small}
\begin{tabular}{cc}
\toprule
Datasets & Number of Inserted \texttt{<Think\_n>} \\
\midrule
WiC & 55 \\
QQP & 38 \\
Winogrande & 32 \\
\bottomrule
\end{tabular}
\label{tab:blank_detail}
\end{small}
\end{center}
\end{table}

\label{app:blank_rationales}
\begin{table}[h]
\label{tab:blank_demo}
\caption{Illustration of blank rationales.}
\begin{center}
\begin{tabular}{|c|m{10cm}|}
\hline
Rationale & The premise provides information about KnowledgeWare, its founders, its headquarters in Atlanta, Georgia, and its product ... The answer is B \\
\hline
Blank Rationale & \texttt{<Think\_0> <Think\_1> <Think\_2> <Think\_3> <Think\_4> <Think\_5> <Think\_6> ... <Think\_n> The answer is B} \\
\hline
\end{tabular}
\label{XXX}
\end{center}
\end{table}

\begin{figure}[h]
    \centering
    \includegraphics[width=0.5\linewidth]{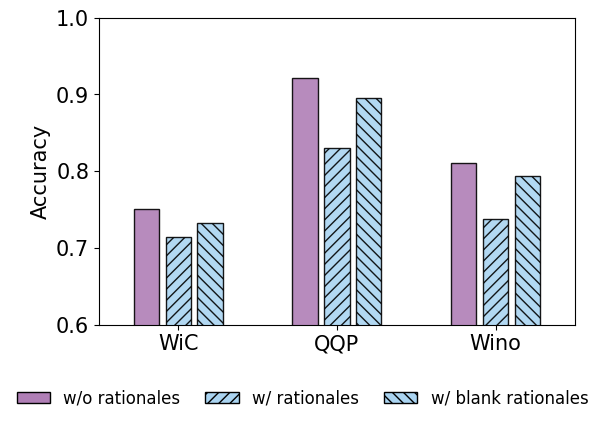}
    \caption{Experimental results of blank rationales.}
    \label{fig:pads}
\end{figure}

Experimental results are shown in Figure~\ref{fig:pads}

\section{Results of Rationale-Augmented Prompting Setting.}
\label{app:rap}
Here are the diagrams that supports the conclusions in RAP setting. 

\begin{figure*}[htb]
\begin{center}
\includegraphics[width=0.9\textwidth]{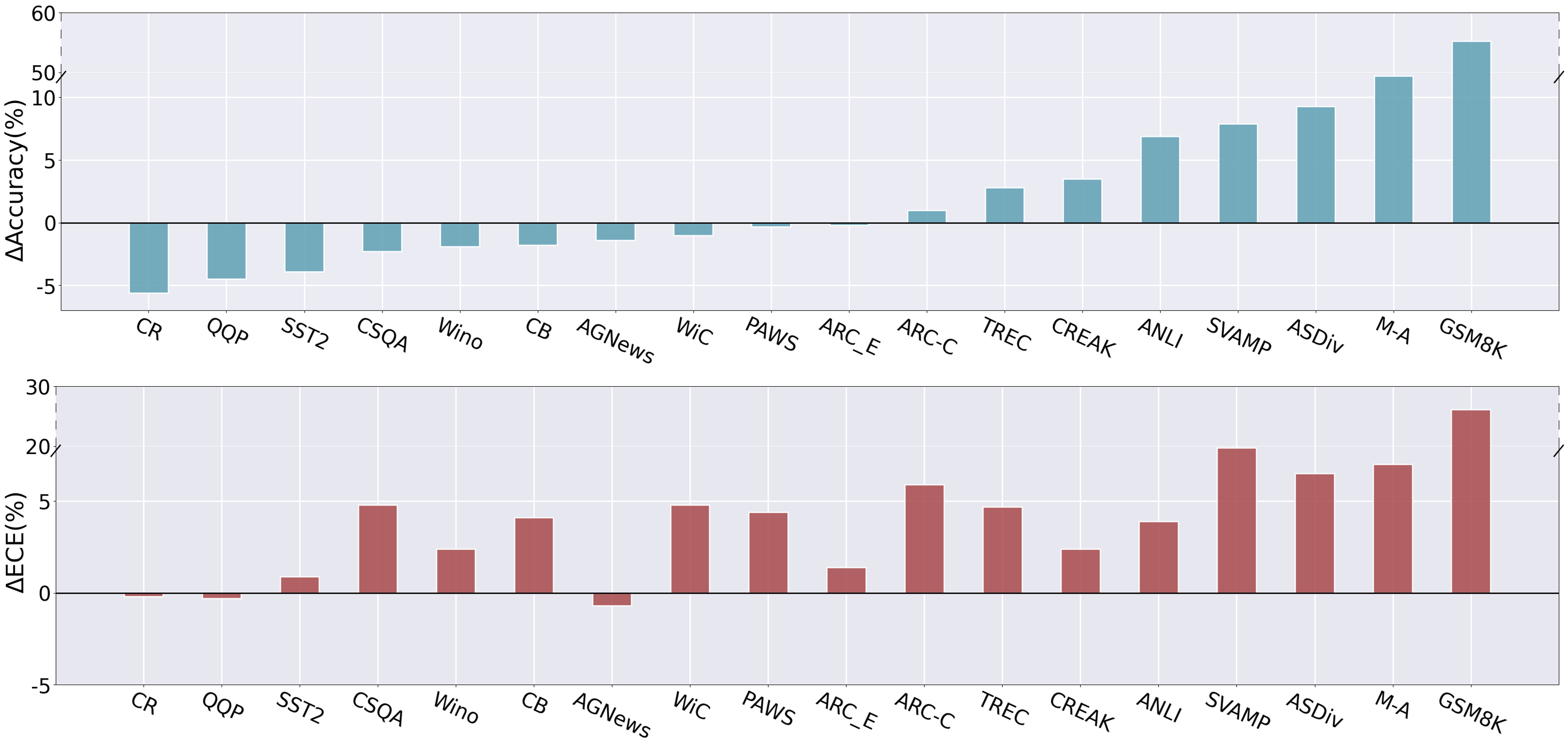}
\caption{Improvements in accuracy and ECE under rationale-augmented prompting setting. Datasets are re-ordered according to the improvements in accuracy. Wino refers to Winogrande and M-A refers to MultiArith.}
\label{fig:bar_chatgpt}
\end{center}
\end{figure*}

\begin{figure*}[ht]
\begin{center}
\includegraphics[width=0.9\textwidth]{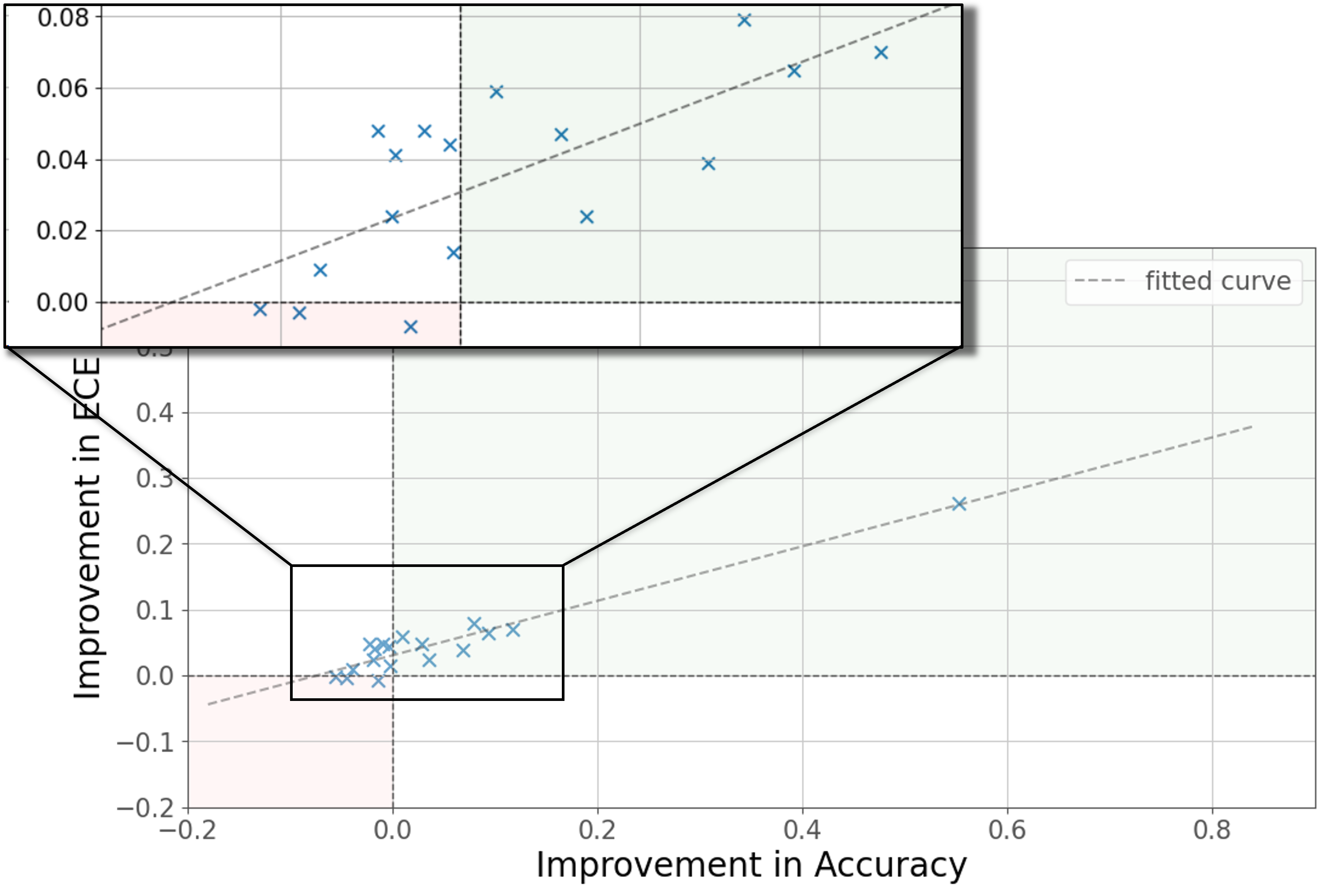}
\caption{Improvement in Accuracy and ECE of \texttt{gpt-3.5-turbo-0613} under rationale-augmented prompting setting. Each point is a datasets, and its x/y-coordinate represents the improvement in model accuracy/ECE respectively. Point labels are omitted as points are close to each other.}
\label{fig:acc_ece_chatgpt}
\end{center}
\end{figure*}

\begin{figure*}[htb]
\begin{center}
\includegraphics[width=\textwidth]{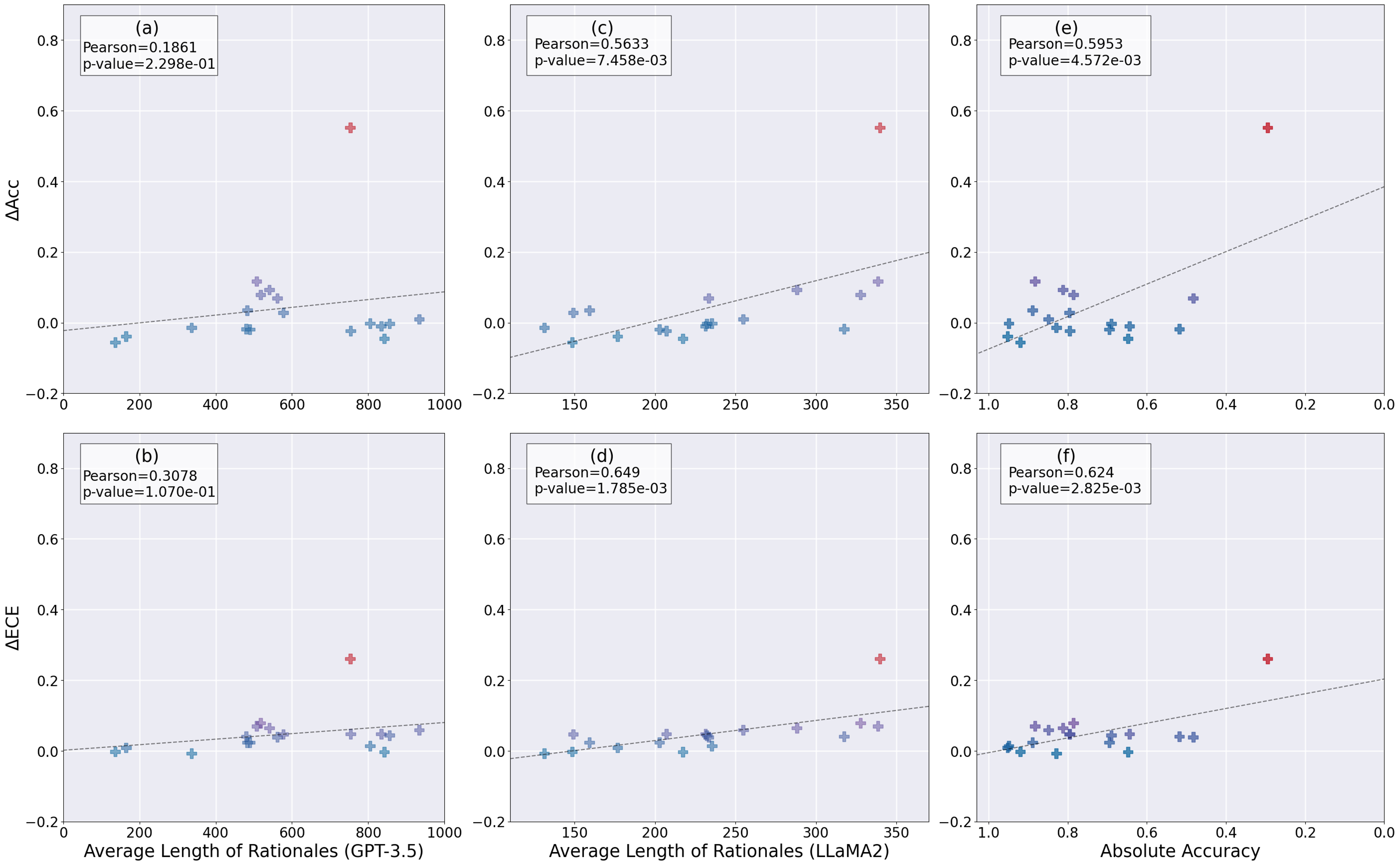}
\caption{Improvement in accuracy and ECE under rationale-augmented prompting setting when task difficulty is measured with different metrics. Subplot (a), (b): Average length of rationales generated by \texttt{gpt-3.5-turbo-0613}. Subplot (c), (d): Average length of rationales generated by LLaMA-2-7B-base. Subplot (e), (f): Original Accuracy of LLaMA-2 model finetuned with answer labels (x-ticks is reversed).}
\label{fig:difficulty_chatgpt}
\end{center}
\end{figure*}

Fig.~\ref{fig:bar_chatgpt} shows the improvement in accuracy and ECE respectively, where the conclusions remain that rationales deteriorate model performance in some tasks, while in most cases benefit model calibration.
Fig.~\ref{fig:acc_ece_chatgpt} and Table \ref{table:linear_fit_chatgpt} shows the relation between improvement in accuracy and ECE. In Fig.~\ref{table:linear_fit_chatgpt} we zoom the dense area and omit point labels for clearance. There are still linear correlation between improvements in model accuracy and ECE, while such correlation may be less visually significant as most points are near the origin.
Fig.~\ref{fig:difficulty_chatgpt} shows the relation between Acc/ECE improvement and different difficulty metrics, in which the positive relation is less pronounced than that in RAFT experiments, which indicates that in RAP setting, model performance and calibration are less affected by task difficulty.

\begin{table*}[h]
\caption{Numeric value of the linear fit between Accuracy and ECE Improvment.}
\label{table:linear_fit_chatgpt}
\begin{center}
\begin{small}
\begin{sc}
\begin{tabular}{lr}
\toprule
Results & Values \\
\midrule
Slope      & 0.413597 \\
Intercept  & 0.030452 \\    
Pearson    & 0.955723 \\
P-Value    & $3.230854e^{-10}$ \\
\bottomrule
\end{tabular}
\end{sc}
\end{small}
\end{center}
\end{table*}

\newpage

\newpage

\begin{table*}[t]
\caption{Pearson and p-Value of the one-tailed hypothesis test for linear relation. Bold entries are p-values lower than 0.05.}
\label{table:linear_fit_comparison}
\begin{center}
\begin{tabular}{cccccc}
\toprule
\multirow{2}{*}{Settings} & \multirow{2}{*}{Metrics} & \multicolumn{2}{c}{Accuracy} & \multicolumn{2}{c}{ECE} \\
\multicolumn{2}{c}{} & Pearson & p-Value & Pearson & p-Value \\
\midrule
\multirow{3}{*}{\textbf{RAFT}} & GPT-3.5 & 0.0010 & 0.4984 & 0.1161 & 0.3232 \\
& LLaMA2  & 0.7059 & $\mathbf{5.3e^{-4}}$ & 0.7802 & $\mathbf{6.7e^{-5}}$ \\
& Accuracy  & 0.9295 & $\mathbf{1.2e^{-8}}$ & 0.9718 & $\mathbf{9.3e^{-12}}$\\
\midrule
\multirow{3}{*}{\textbf{RAP}} & GPT-3.5 & 0.1861 & 0.2298 & 0.3078 & 0.1070 \\
& LLaMA2  & 0.5633 & \textbf{0.0075} & 0.649 & \textbf{0.0018} \\
& Accuracy  & 0.5953 & \textbf{0.0046} & 0.6240 & \textbf{0.0028} \\
\bottomrule
\end{tabular}
\end{center}
\end{table*}

\clearpage
\section{Results of Multi-task Training}
\label{app:multi_task}
\begin{figure*}[h]
\centering
\includegraphics[width=0.9\textwidth]{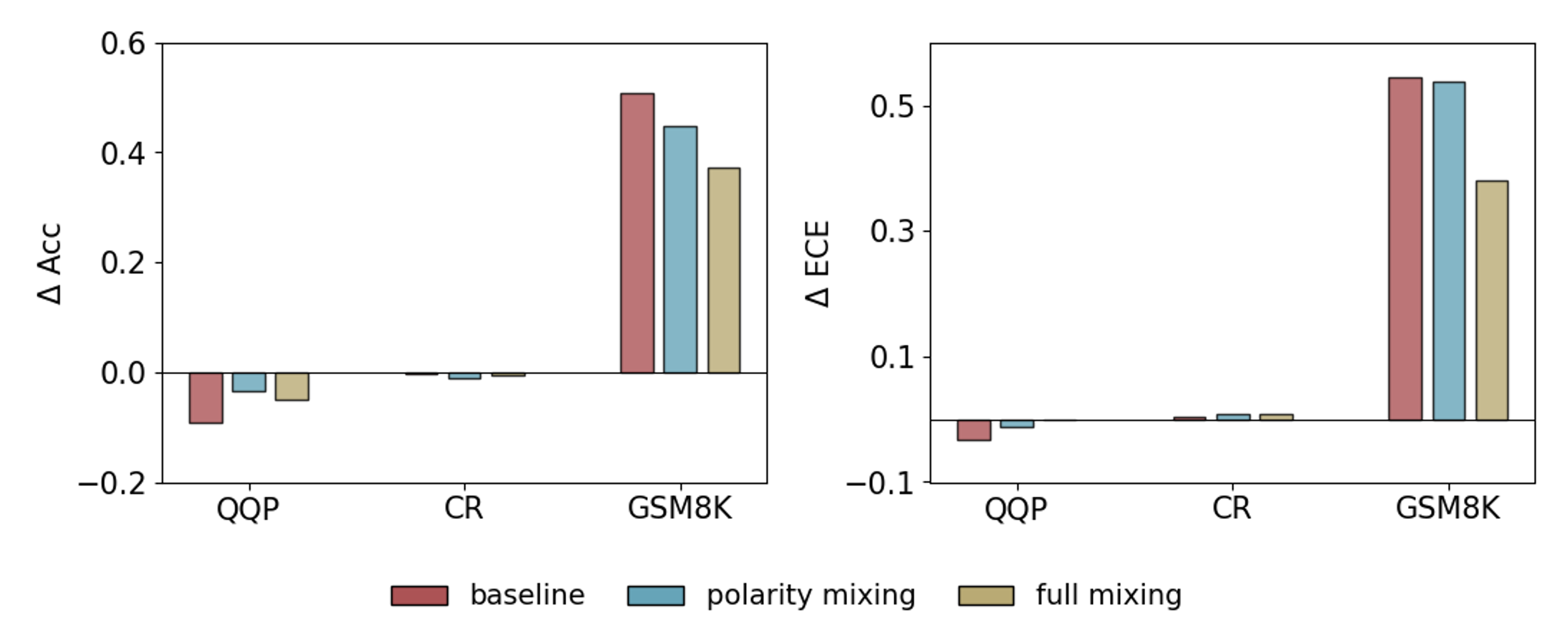}
\caption{Improvement in Accuracy and ECE of models trained in multi-task settings. (a) Baseline: single task finetuning. (b) Polarity mixture: mix data from datasets where RAFT cause performance increase (or drop). (c) Full mixture: mix data from all datasets.}
\label{fig:multi_task}
\end{figure*}

Fig.~\ref{fig:multi_task} shows the improvement in accuracy and ECE of models trained in multi-task settings. In polarity mixture setting, for data where RAFT brings gain, we mix QQP, Winogrande, and CR. For the other one, we mix GSM8K, ARC, and CREAK. All results under multi-task learning setting are on the same side of y-axis with baselines, which indicates that multi-task instruction tuning does not change whether RAFT acts positively or negatively.

\end{document}